\documentclass[fleqn,10pt,twocolumn]{wlscirep}
\usepackage[utf8]{inputenc}
\usepackage[T1]{fontenc}
 \usepackage{booktabs}
 \usepackage{multirow}
 \usepackage{graphicx}
 \usepackage{graphics}
 \usepackage{caption}
\usepackage{subcaption}
\usepackage{color,soul}
\usepackage{xcolor}

\usepackage[switch]{lineno}

\newcommand{\revised}[1]{\textcolor{black}{#1}}

\newcommand{\npj}[1]{\textcolor{black}{#1}}

\title{Longitudinal cardio-respiratory fitness prediction through wearables in free-living environments}

\author[1,*+]{Dimitris Spathis}
\author[2+]{Ignacio Perez-Pozuelo}
\author[2]{Tomas I. Gonzales}
\author[1]{Yu Wu}
\author[2]{Soren Brage}
\author[2]{Nicholas Wareham}
\author[1]{Cecilia Mascolo}
\affil[1]{Department of Computer Science and Technology, University of Cambridge, UK}
\affil[2]{MRC Epidemiology Unit, School of Clinical Medicine, University of Cambridge, UK}

\affil[*]{corresponding author: ds806@cl.cam.ac.uk}

\affil[+]{these authors contributed equally to this work}

\begin{abstract}

Cardiorespiratory fitness is an established predictor of metabolic disease and mortality. Fitness is directly measured as maximal oxygen consumption (VO$_{2}max$), or indirectly assessed using heart rate responses to standard exercise tests. However, such testing is costly and burdensome because it requires specialized equipment such as treadmills and oxygen masks, limiting its utility. Modern wearables capture dynamic real-world data which could improve fitness prediction. In this work, we design algorithms and models that convert raw wearable sensor data into cardiorespiratory fitness estimates. We validate these estimates' ability to capture fitness profiles in free-living conditions using the Fenland Study (N=11,059), along with its longitudinal cohort (N=2,675), and a third external cohort using the UK Biobank Validation Study (N=181) who underwent maximal VO$_{2}max$ testing, the gold standard measurement of fitness. Our results show that the combination of wearables and other biomarkers \npj{as inputs to neural networks} yields a strong correlation to ground truth in a holdout sample ($r$ = 0.82, 95CI 0.80-0.83), \npj{outperforming other approaches and models} and detects fitness change over time (e.g., after 7 years). \npj{We also show how the model's latent space can be used for fitness-aware patient subtyping paving the way to scalable interventions and personalized trial recruitment}. These results demonstrate the value of wearables for fitness estimation that today can be measured only with laboratory tests.

\end{abstract}
\begin{document}

\flushbottom
\maketitle
%
%
\thispagestyle{empty}


\section*{Introduction}
Cardiorespiratory fitness (CRF) is one of the strongest known predictors of cardiovascular disease (CVD) risk and is inversely associated with many  other health outcomes ~\cite{mandsager2018association}. CRF is also a potentially stronger predictor of CVD outcomes when compared to other risk factors like hypertension, type 2 diabetes, high cholesterol, and smoking.  Despite its prognostic value, routine CRF assessment remains uncommon in clinical settings because maximal oxygen consumption (VO$_{2}max$), the  \textit{gold-standard} measure of CRF, is challenging to directly measure. A computerised gas analysis system is needed to monitor ventilation and expired gas fractions during exhaustive exercise on a treadmill or cycle ergometer. Additional equipment may be needed to monitor other biosignals, such as heart rate (HR). These equipment require trained research personnel to operate, and an attending physician is a requisite for exercise testing in some scenarios. Several criteria must also be achieved to verify that exhaustion has been reached, including leveling-off of VO$_{2}$, achieving a percentage of age-predicted maximal HR, and surpassing a peak respiratory exchange ratio threshold \cite{swain2014acsm}. The costs of VO$_{2}$ measurement and risks of exhaustive exercise not only limit direct CRF assessment in clinical settings, but also restrict research of CRF at the population level. Thus, our understanding of differences in CRF within populations, across geographic regions, and over time is lacking.

Non-exercise prediction models of VO$_{2}max$ are an alternative to exercise testing in clinical settings. These models are usually regression based and incorporate variables like sex, age, body mass index (BMI), resting heart rate (RHR), and self-reported physical activity ~\cite{cao2010predicting}. We have recently shown that RHR alone can be used to estimate VO$_{2}max$~\cite{gonzales2020resting}, however the validity of estimates from this approach are considerably lower than those achieved with  exercise testing~\cite{gonzales2020estimating,nes2011estimating}. \npj{Also, the response of heart rate to activity has been shown to be predictive of VO$_{2}max$, in coarse-grained data \cite{spathis2021self}}. Wearable devices such as activity trackers and smartwatches can monitor not only RHR and physical activity but other biosignals in free-living conditions \cite{dunn2021wearable}, potentially enabling more precise estimation  of VO$_{2}max$ without exercise testing.
Recent attempts to use wearable devices to estimate VO$_{2}max$ are difficult to externally evaluate, however, because their estimation methods tend to be non-transparent ~\cite{shcherbina2017accuracy} and lack scientific validation  ~\cite{passler2019validity,shcherbina2017accuracy}. Although certain wearable devices show promise, they tend to rely on detailed physical activity intensity measurements, GPS-based speed monitoring, and require users to reach near-maximal HR values, which limits their use to fitter individuals~\cite{cooper2019validity}. \npj{Some studies attempt to estimate VO$_{2}max$ from data collected during free-living conditions, but these are typically from small-scale cohorts and use contextual data from treadmill activity, which restricts their application in population settings \cite{altini2016cardiorespiratory, helgerud2022prediction}}.

\begin{figure*}
    \centering
    \includegraphics[width=1.\linewidth]{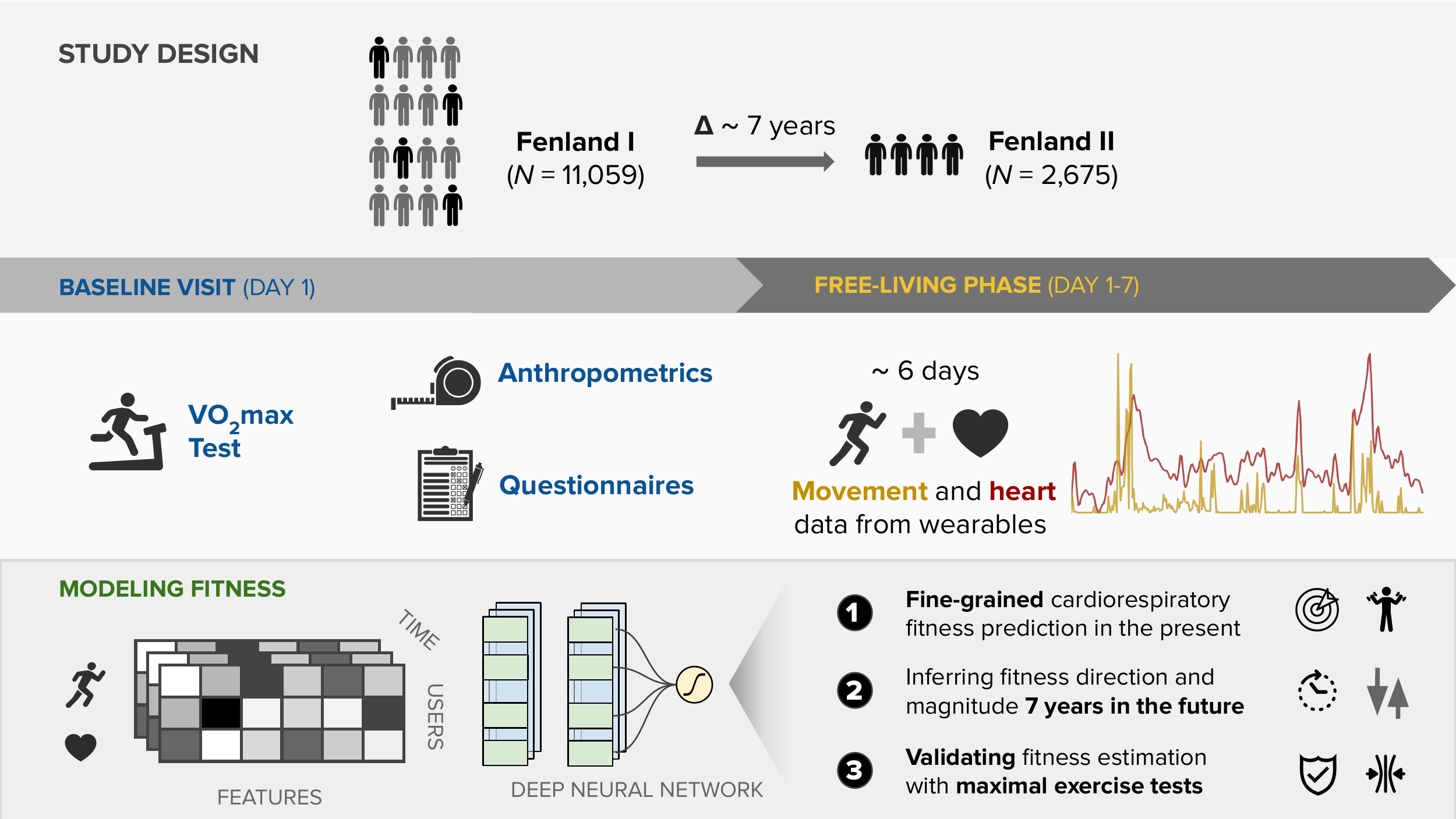}
	\caption{\npj{\textbf{Study and experimental design}. We use a cohort study of 11,059 participants with laboratory and wearable sensor data and a longitudinal subsample of 2,675  participants who were repeated the protocol 7 years later. Using the free-living sensors as input data, we train machine learning models to predict lab-measured cardio-respiratory fitness (VO2max).   }}
    \label{fig:study_hero} 
\end{figure*}



Here, we use data from the largest study of its kind, by over two orders of magnitude, and use purely free-living data to predict VO$_{2}max$, with no requirement for context-awareness. This work substantially advances previous non-exercise models for predicting CRF by introducing an adaptive representation learning approach to physiological signals derived from wearable sensors in a large-scale population with free-living condition data. We employ a deep neural network model that utilizes feedforward non-linear layers to learn personalized fitness representations.
We demonstrate that these models yield better performance than traditional and state-of-the-art non-exercise models. We illustrate how these models can be used to predict the magnitude and direction of change in CRF. Moreover, we show that they can adapt in time given behavioural changes by showcasing strong performance in a subset of the same population who were retested seven years later. This has implications for the estimation of  population fitness levels and lifestyle trends, including the identification of sub-populations or areas in particular need of intervention. Such models can improve both population-health and personalized medicine applications. For instance, the ability of the patient to undergo certain treatments like surgery \cite{richardson2017fit} or chemotherapy \cite{leon2017fit} can be assessed through wearables before the procedure takes place and therefore reduce postoperative complications.




\section*{Results}

\begin{figure*}
\centering

\medskip
 \includegraphics[width=1.\linewidth]{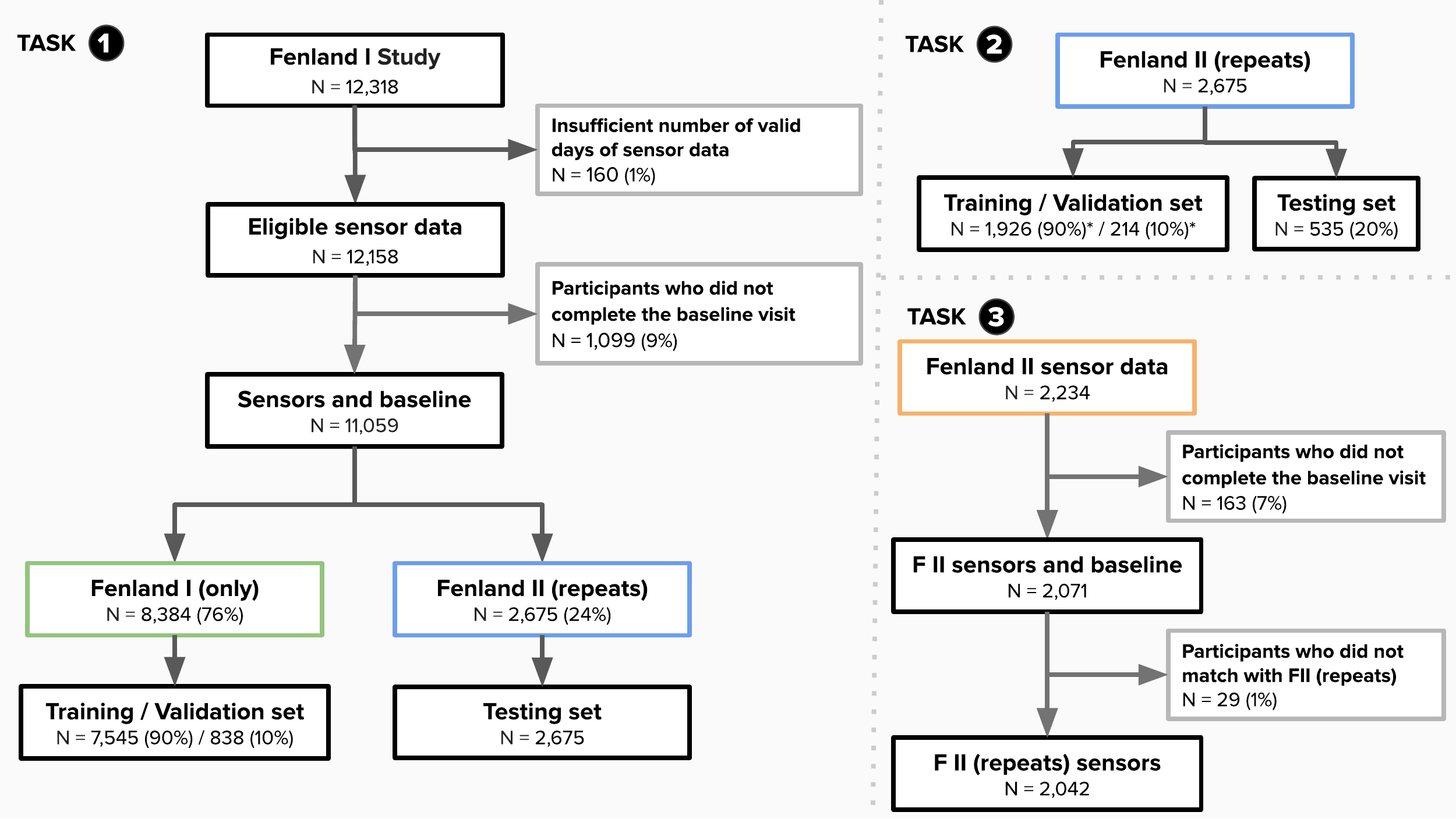}

\bigskip
\resizebox{1\textwidth}{!}{
\begin{tabular}{lllllllllllll}
\toprule
 & \multicolumn{4}{c}{\textbf{Fenland$_{present}$  }} &
 \multicolumn{4}{c}{\textbf{Fenland II$_{future}$ }} &
 \multicolumn{4}{c}{\textbf{Fenland II$_{present}$ }}
 \\\cmidrule(lr){2-5}\cmidrule(lr){6-9}\cmidrule(lr){10-13}
 & \multicolumn{2}{l}{Men (n= 5229)} & \multicolumn{2}{l}{Women (n= 5830)} & \multicolumn{2}{l}{Men (n=1303)} & \multicolumn{2}{l}{Women (n=1372)} & \multicolumn{2}{l}{Men (n=1303)} & \multicolumn{2}{l}{Women (n=1372)}
 \\\cmidrule(lr){2-3}\cmidrule(lr){4-5}\cmidrule(lr){6-7}\cmidrule(lr){8-9}\cmidrule(lr){10-11}\cmidrule(lr){12-13}
 & \textit{mean}  & \textit{std}  & \textit{mean}  & \textit{std}  & \textit{mean}  & \textit{std}  & \textit{mean}  & \textit{std} & \textit{mean}  & \textit{std} & \textit{mean}  & \textit{std} \tabularnewline
 \midrule
\multicolumn{13}{l}{\textbf{Demographics} }\tabularnewline
\qquad{}Age (years)  & 47.70 & 7.57 & 47.66 & 7.36 & 54.11 & 7.08 & 54.76 & 6.81 & 47.19 & 7.18 & 47.83 & 6.96  \tabularnewline
\multicolumn{13}{l}{\textbf{Anthropometrics} }\tabularnewline
\qquad{}Height (m)  & 1.78 & 0.07 & 1.64 & 0.06 & 1.77 & 0.06 & 1.64 & 0.06 & 1.77 & 0.06 & 1.64 & 0.06\tabularnewline
\qquad{}Body mass (kg)  & 85.85 & 13.83 & 70.54 & 13.92 & 85.31 & 13.59 & 69.58 & 13.77 & 84.85 & 13.14 & 69.04 & 13.26 \tabularnewline
\qquad{}BMI (kg/m$^{2}$)  & 27.16 & 3.97 & 26.17 & 4.97 & 27.03 & 4.01 & 25.85 & 4.94 & 27.00 & 4.03 & 25.84 & 4.98 \tabularnewline

\multicolumn{13}{l}{\textbf{Physical activity} }\tabularnewline
\qquad{}MVPA (min/day)  & 35.87 & 22.35 & 34.40 & 22.59 & 34.92* & 22.18* & 35.35*  & 23.26* & 34.41 & 22.23 & 32.81 & 21.45 \tabularnewline
\qquad{}VPA (min/day)  & 3.27 & 8.57 & 3.31 & 15.67 & 3.57*  & 8.78*  & 3.30*  & 7.52* & 3.38 &9.30 & 3.86 & 27.80 \tabularnewline

\multicolumn{13}{l}{\textbf{Resting Heart Rate} }\tabularnewline
\qquad{}RHR (bpm)  & 61.48 & 8.68 & 64.46 & 8.28 & 59.63*  & 8.28*  & 62.21* & 8.10* & 61.06 & 8.44 & 63.81 & 8.20 \tabularnewline

\multicolumn{13}{l}{\textbf{Cardiorespiratory fitness} }\tabularnewline
\qquad{}VO$_{2}$max (ml $O_2/min/kg$)  & 41.95 & 4.61 & 37.44 & 4.73  & 42.32 & 4.68  & 37.93  & 4.72 & 42.21 & 4.42 & 37.84 & 4.69 \tabularnewline

\bottomrule
\tabularnewline
\end{tabular}
}

	\caption{\textbf{Characteristics of the study analytical sample across the three tasks.} \textbf{Top:} The first task trains a model to predict fitness using the large cohort (Fenland I), the second task is using the smaller cohort of repeats in Fenland I (called Fenland II) and trains further models to predict fitness now and in the future (and their delta). The third task evaluates the original model trained in Task 1 by feeding new sensor data to assess the adaptability of the model to pick up change. (*Training set is 90\% of the 80\% remaining dataset after splitting to testing set. Validation set is 10\% of the training set). \textbf{Bottom}: Dataset statistics breakdown by sex and features [data is in mean (std)]. Values with asterisk(*) indicate that this variable comes from Fenland II sensor data which is a smaller cohort (N=2071) due to data filtering (see Top panel, Task 3). The values in FII (future) cohort correspond to the second assessment (7 years later).}  
   
 \label{fig:flowchart} 
\end{figure*}

Baseline measurements were collected from 12,435 healthy adults from the Fenland study in the United Kingdom~\cite{lindsay2019descriptive}, where all required data for the present analysis were available in 11,059 participants (Fenland I, baseline timepoint referred to as "current" in our evaluation). A subset of 2,675 participants were assessed again after a median (interquantile range) of 7 (5-8)  years (Fenland II, referred to as "future" in our evaluations). Descriptive characteristics of the two analysis samples are presented in Figure~\ref{fig:flowchart}. \revised{ We present the characteristics of the longitudinal cohort in both temporal snapshots in  Figure~\ref{fig:flowchart} ("present" and "future").} Mean and standard deviations for each characteristic are presented in this table. An overview of the study design and the three experimental tasks is provided in Figure~\ref{fig:study_hero}.

\subsection*{Fine-grained fitness prediction from wearable sensors}

\begin{figure*} 

\begin{subfigure}{0.55\textwidth}
\includegraphics[scale=0.45]{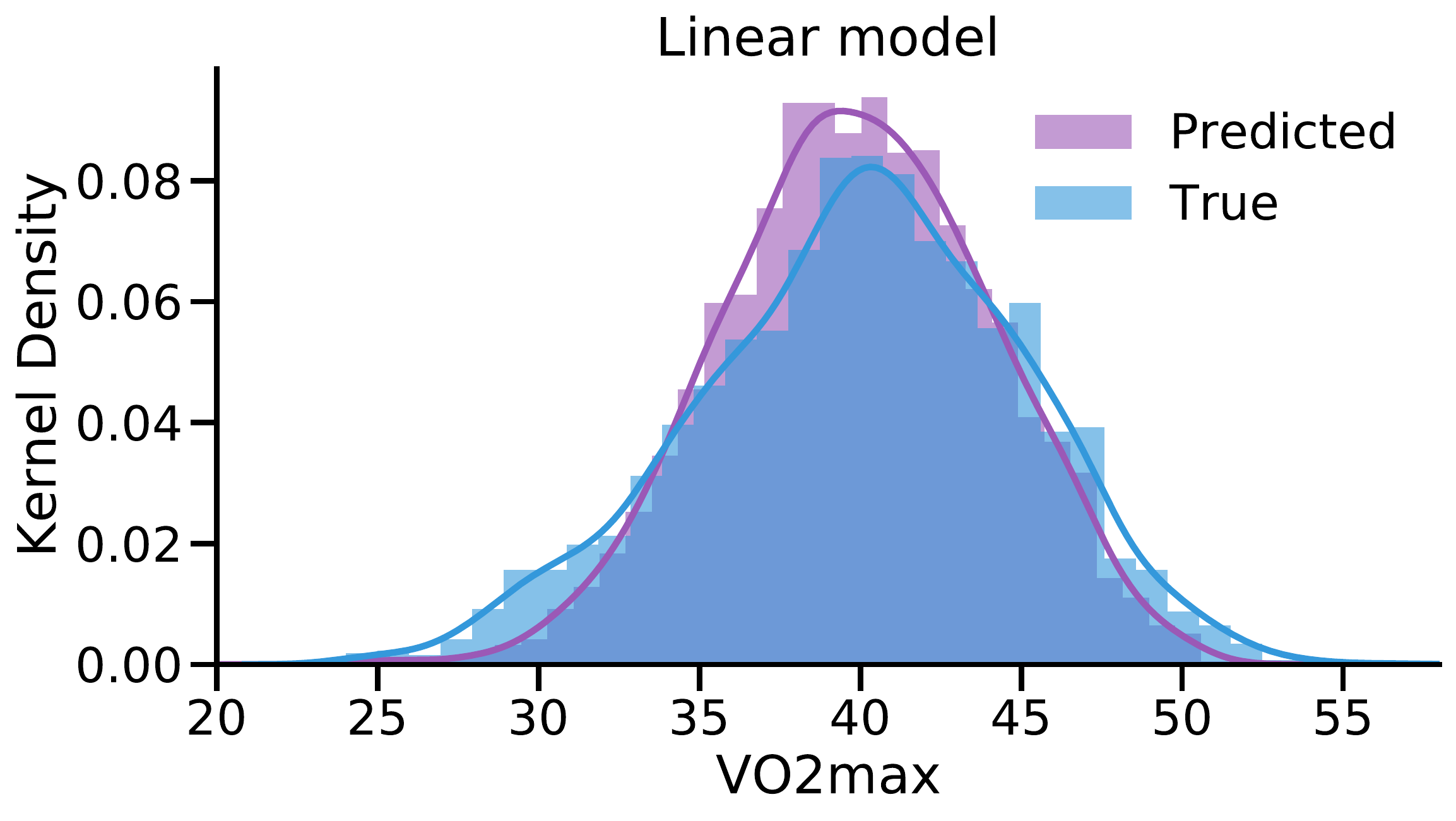}
\caption{} \label{fig:FI_evaluation:a}
\end{subfigure}\hspace*{\fill}
\hspace{10pt}
\begin{subfigure}{0.35\textwidth}
\includegraphics[scale=0.4]{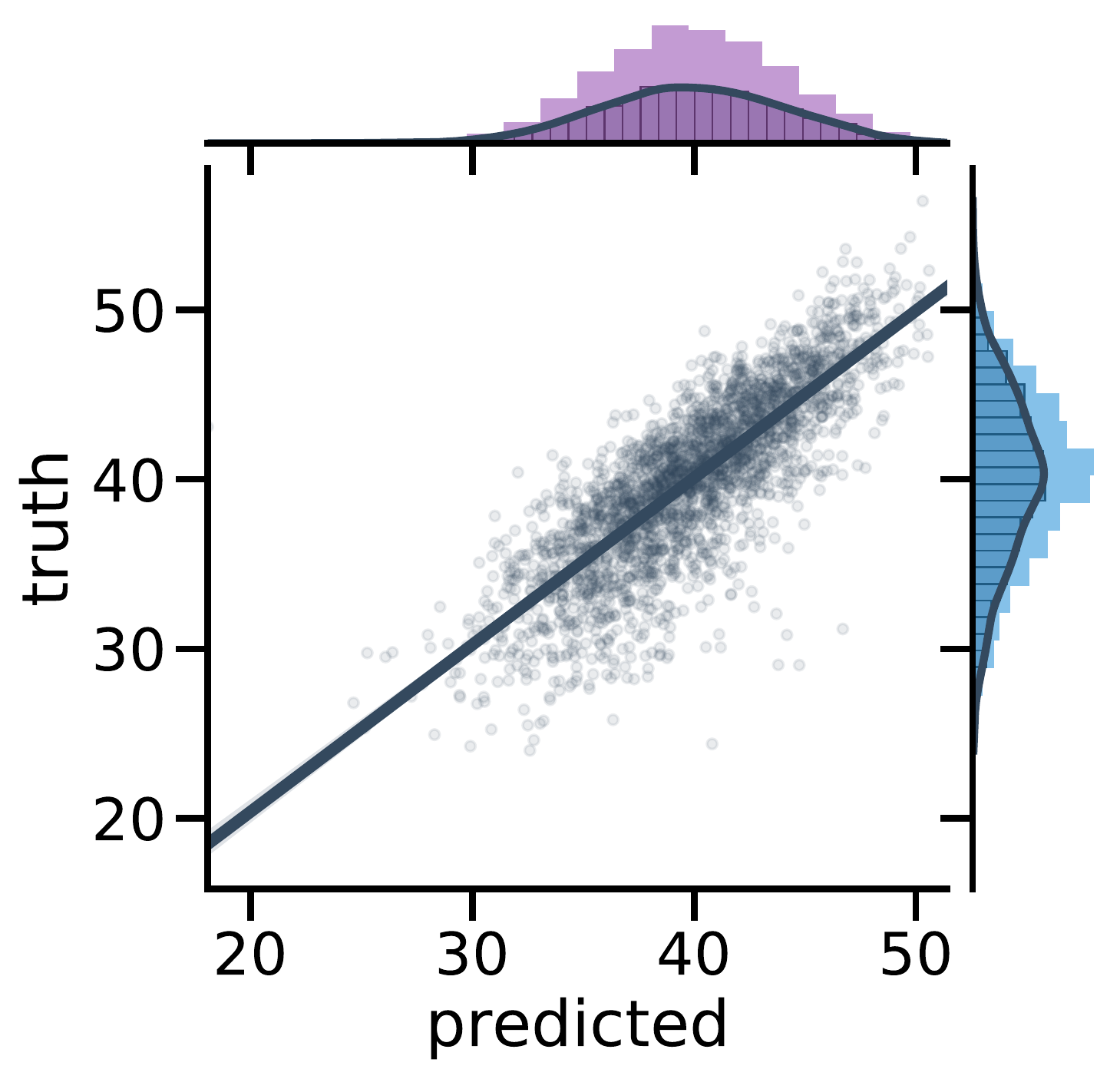}
\caption{} \label{fig:FI_evaluation:b}
\end{subfigure}

\medskip
\begin{subfigure}{0.55\textwidth}
\includegraphics[scale=0.45]{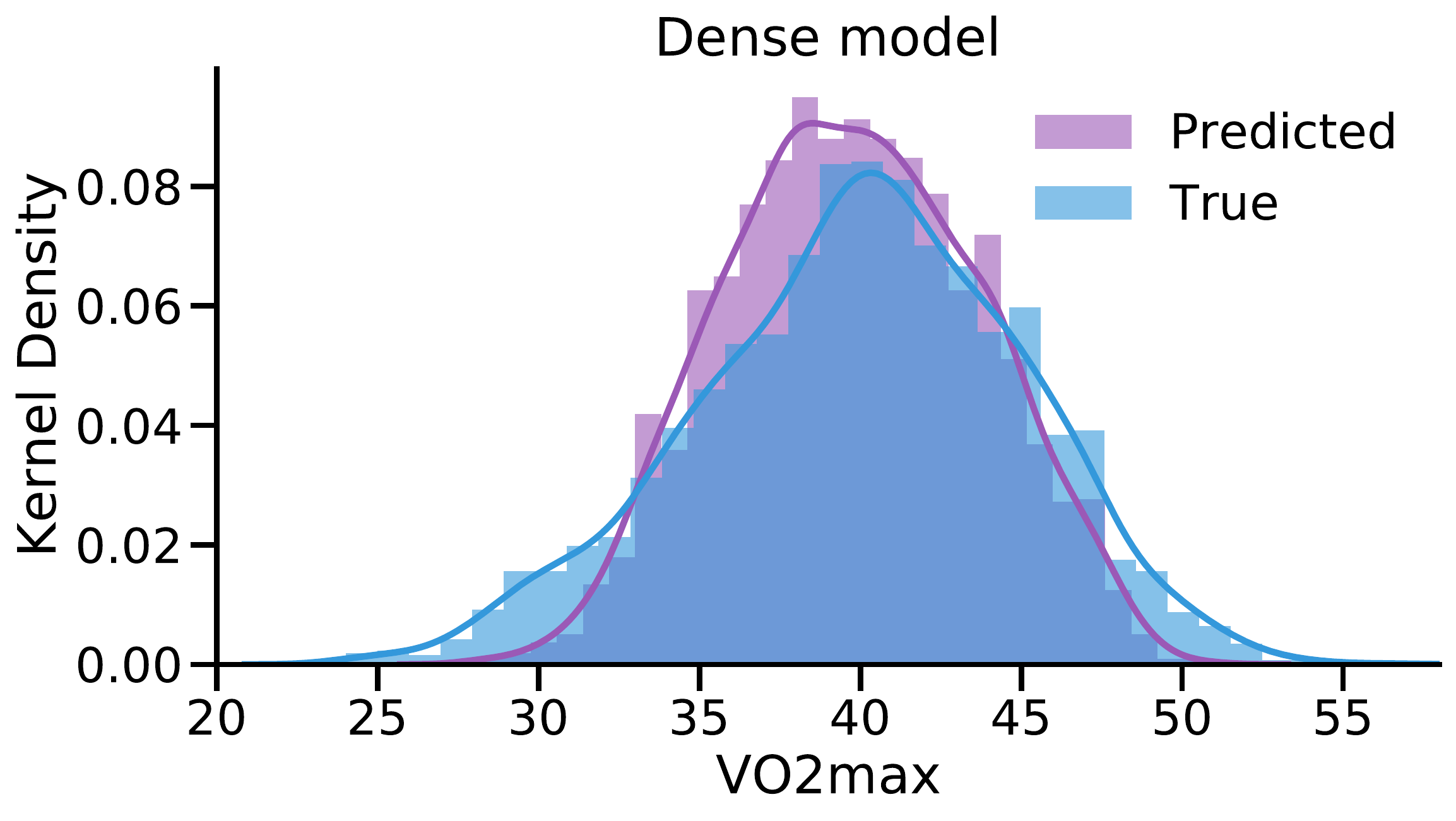}
\caption{} \label{fig:FI_evaluation:c}
\end{subfigure}\hspace*{\fill}
\hspace{10pt}
\begin{subfigure}{0.35\textwidth}
\includegraphics[scale=0.4]{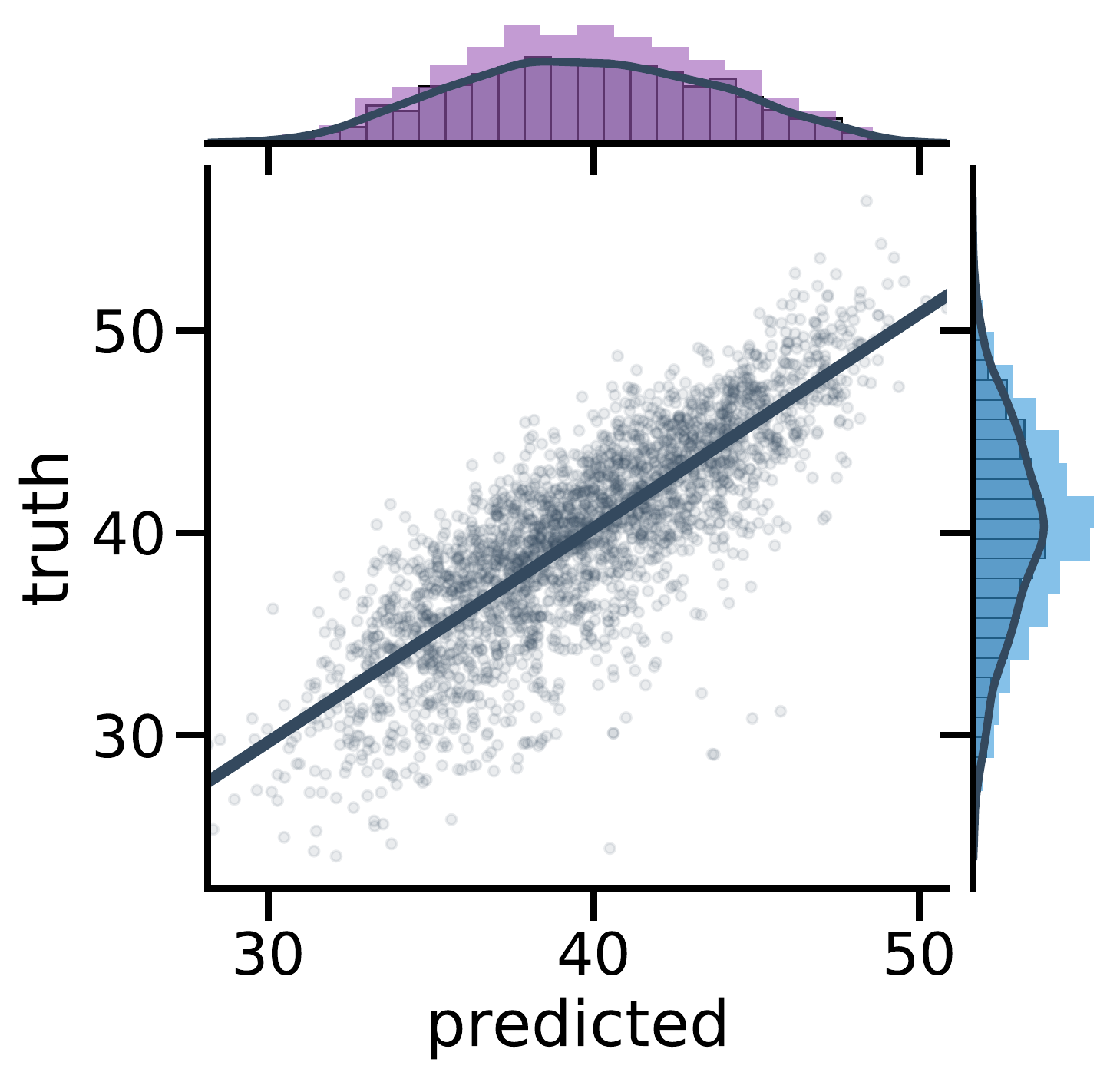}
\caption{} \label{fig:FI_evaluation:d}
\end{subfigure}


\caption{\npj{\textbf{Comparison of fine-grained fitness prediction with the two comprehensive models}. Comparing the predicted and true VO2max coming from the best performing comprehensive model (\textit{Sensors + RHR + Anthro.}) trained with Fenland I. \textbf{(a, c)}  Linear and dense models produce accurate predictions with correlation of predicted and true VO2max up to $r=0.82$, $p<0.005$ (see Table~\ref{tab:fenI_results}) \textbf{(b, d, )}. The plot combines a kernel density estimate and histogram, while the gray line denotes a linear regression fit. Transparency has been applied to the datapoints to combat crowding.}}
    
\label{fig:FI_evaluation}
\end{figure*}

We first developed and externally validated several non-exercise VO$_{2}$max estimation models as a regression task using features commonly measured by wearable devices (anthropometry, resting heart rate (RHR), physical activity (PA); see Table~\ref{tab:fenI_results}). Here our goal was to explore how conventional non-exercise approaches to VO$_{2}$max estimation could be enhanced by features from free-living PA data. We split participant data into independent training and test sets. The training set (n=8384, participants with baseline data only) was used for model development. The test set (n=2675, participants with baseline and followup data) was used to externally validate each model. \npj{Starting with linear regression models, we used anthropometry or RHR alone which yielded poor external validity (R${^2}$ of 0.35-37), but validity improved when combined in the same model (R${^2}$ of 0.61). The best performance (R${^2}$ of 0.67) was attained using a deep neural network model combining wearable sensors, RHR, and anthropometric data (Figure~\ref{fig:FI_evaluation})}. For reference, we compare these results to traditional non-model equations, which rely on Body Mass, RHR, and Age. Using a popular equation (as proposed in \cite{uth2004estimation, tanaka2001age}) we obtained poor validity (R${^2}$ of -3.2 and 
Correlation of 0.389), a  performance lower than  using anthropometrics only in our setup (see Methods for details). This motivates the use of machine learning which captures better covariate interactions.

\npj{To understand the limits of the models, we conducted a number of post-hoc sensitivity analyses by investigating subgroup performance in terms of sex (male/female), age, weight, BMI, and height (Table \ref{tab:breakdown}), as well as investigating model errors on Bland-Altman agreement plots (Figure \ref{fig:bland_alt}). We found that the comprehensive model is robust to most subgroups, showing minimal differences in most cases, with exceptions in weight and age. In particular, we found no difference between male and female participants even though the performance was lower for each group compared to the mixed set (R${^2}$ of 0.59). Further, the models perform better on participants of lower age (R${^2}$ of 0.68), higher weight (R${^2}$ of 0.69). No effects were observed on height or BMI differences (overlapping CIs). The best performing subgroup is "higher weight" and the worst performing "sex-male". Last, Bland-Altman plots showed that the Dense model has better upper difference compared to the linear model, where the lower and mean difference were similar (Figure \ref{fig:bland_alt}). }

\npj{Deep neural networks can learn feature representations that are suitable for clustering tasks, such as population stratification by implicit health status, but are difficult to reveal using linear dimension-reduction techniques \cite{gaspar2019probabilistic}. We used t-distributed stochastic neighbor embedding (tSNE), a nonlinear dimension-reduction technique, to visualise learned feature representations from our model and their relationship to participant VO$_{2}$max (Figure~\ref{fig:tsne}). Clustering and coloring
by VO$_{2}$max was shown to be inversely related and more apparent in
the learned latent space compared to the original feature space. Further, we show how this latent space can be used for patient subtyping through embedding neighbours. Starting from a initial patient (query), we retrieved the five nearest neighbours in the latent and original space. In a case study with three randomly selected participants, we found that the total euclidean distance of the query to all neighbours is higher in the original than the latent space, pointing to better semantic clustering (Figure~\ref{fig:tsne}, bottom panel). 
}


\begin{table*}[]
\caption{\npj{\textbf{Model comparison for predicting fine-grained VO2max with the Fenland I cohort}. Comparison between linear regression and a dense neural network trained on combinations of antrhopometrics, common biomarkers (RHR), and passively collected data over a week (wearable sensors). Best performance in bold. The units of VO2max are measured in  $ml O_2/min/kg$. Results reported from the testing set.}}

\resizebox{0.99\textwidth}{!}{
\begin{tabular}{llllllc}
\hline
\multicolumn{2}{c}{\multirow{2}{*}{Features}}                                                                                 & \multirow{2}{*}{Models}          & \multicolumn{3}{c}{Evaluation Metrics {[}95\% CI{]}}                                                  & \multicolumn{1}{l}{N (train/test set)}                                         \\ \cline{4-7} 
\multicolumn{2}{c}{}                                                                                                          &                                  & \multicolumn{1}{c}{R2}           & \multicolumn{1}{c}{Corr}         & \multicolumn{1}{c}{RMSE}        & \multirow{11}{*}{\begin{tabular}[c]{@{}c@{}}11059 \\ (8384/2675)\end{tabular}} \\ \cline{1-6}
\multicolumn{2}{l}{\textbf{Anthropometrics}}                                                                                  & \multirow{8}{*}{\textbf{Linear}} &                                  &                                  &                                 &                                                                                \\
          & Age/Sex/Weight/BMI/Height                                                                                         &                                  & 0.359 {[}0.329-0.388{]}          & 0.600 {[}0.577-0.623{]}          & 4.051 {[}3.947-4.170{]}         &                                                                                \\
\multicolumn{2}{l}{\textbf{Resting Heart Rate}}                                                                               &                                  &                                  &                                  &                                 &                                                                                \\
          & RHR (sensor-derived)                                                                                              &                                  & 0.373 {[}0.342-0.403{]}          & 0.612 {[}0.587-0.638{]}          & 4.007 {[}3.885-4.113{]}         &                                                                                \\
\multicolumn{2}{l}{\textbf{Anthropometrics + RHR}}                                                                            &                                  &                                  &                                  &                                 &                                                                                \\
          & Age/Sex/Weight/BMI/Height/RHR                                                                                     &                                  & 0.610 {[}0.582-0.634{]}          & 0.781 {[}0.764-0.796{]}          & 3.159 {[}3.051-3.272{]}         &                                                                                \\
\multicolumn{2}{l}{\textbf{Wearable Sensors + RHR + Anthro.}}                                                                 &                                  &                                  &                                  &                                 &                                                                                \\
          & \multirow{2}{*}{\begin{tabular}[c]{@{}l@{}}Acceleration/HR/HRV/MVPA\\ Age/Sex/Weight/BMI/Height/RHR\end{tabular}} &                                  & 0.658 {[}0.623-0.685{]}          & 0.812 {[}0.792-0.828{]}          & 2.956 {[}2.830-3.082{]}         &                                                                                \\ \cline{3-6}
\textbf{} &                                                                                                                   & \textbf{Dense}                   & \textbf{0.671 {[}0.649-0.691{]}}          & \textbf{0.821 {[}0.806-0.835{]}}          & \textbf{2.902 {[}2.806-3.002{]}}         &                                                                                                                                                            \\ \hline
\end{tabular}
\label{tab:fenI_results}
}
\end{table*}

\subsection*{Predicting magnitude and direction of fitness change in the future}


The second group of tasks evaluated our model on the subset of participants who returned for Fenland II $\approx$ 7 years later (referred to as \textit{future} in our evaluations). For these experiments we carried out three evaluations. Following the process described earlier, we re-trained a model to predict future VO$_{2}max$ using only information from the present as input (Table~\ref{tab:fenII_results}). This model yielded a slightly lower accuracy than Fenland I, achieving a R${^2}$ of 0.49 and a correlation of 0.72. This lower performance is expected since the model has no information of the behavior of the individuals 7 years later. We also trained a model to directly predict the difference (or delta) of current-future VO$_{2}max$, which reached a correlation of 0.23.

\begin{figure*}
    \begin{subfigure}{.5\textwidth}
        \centering
        \includegraphics[scale=0.43]{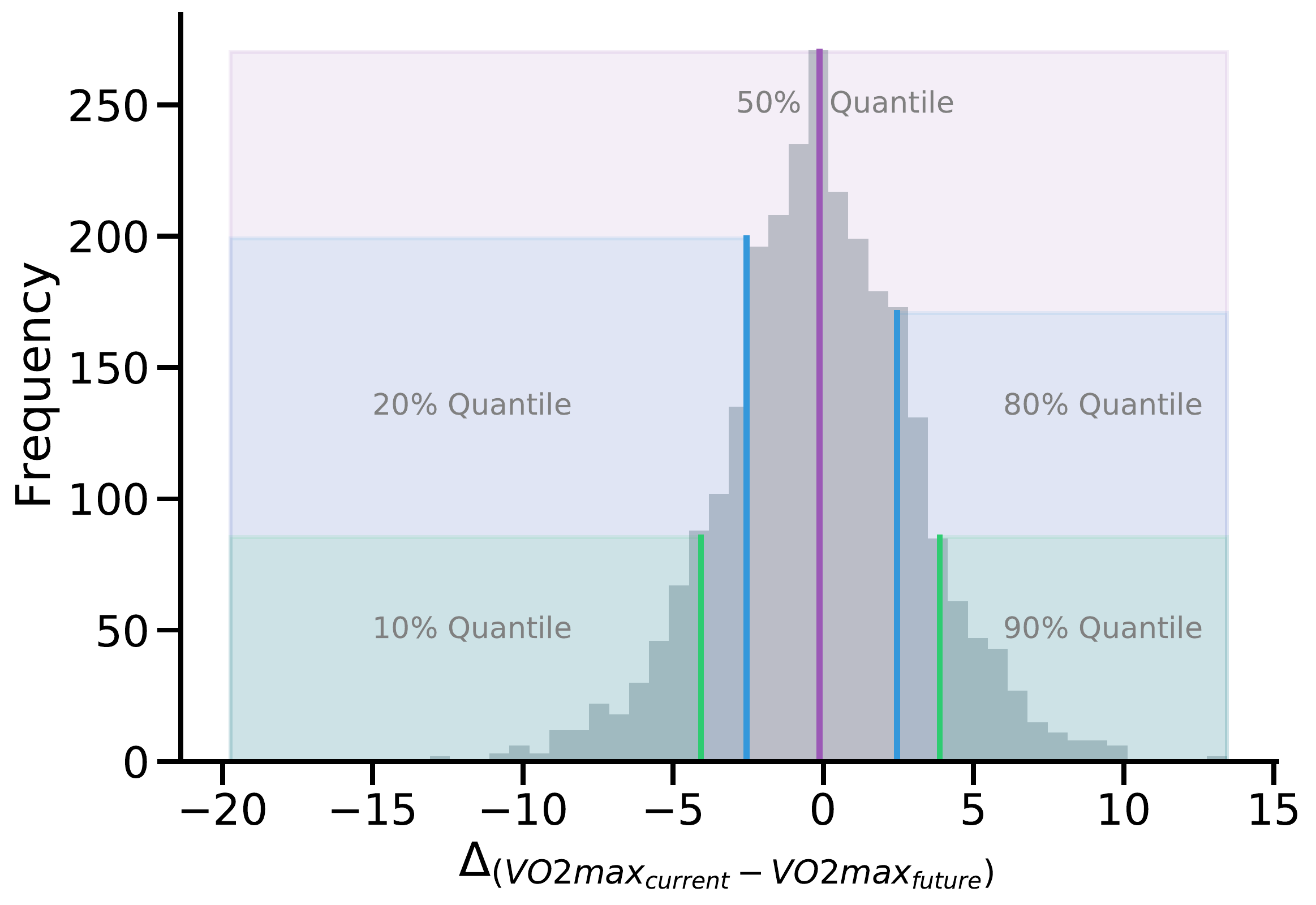}
        \caption{}
        \label{fig:FII_binary:a}
    \end{subfigure}
   \hspace{10pt}
    \begin{subfigure}{.5\textwidth}
        \centering
        \includegraphics[scale=0.38, angle=0]{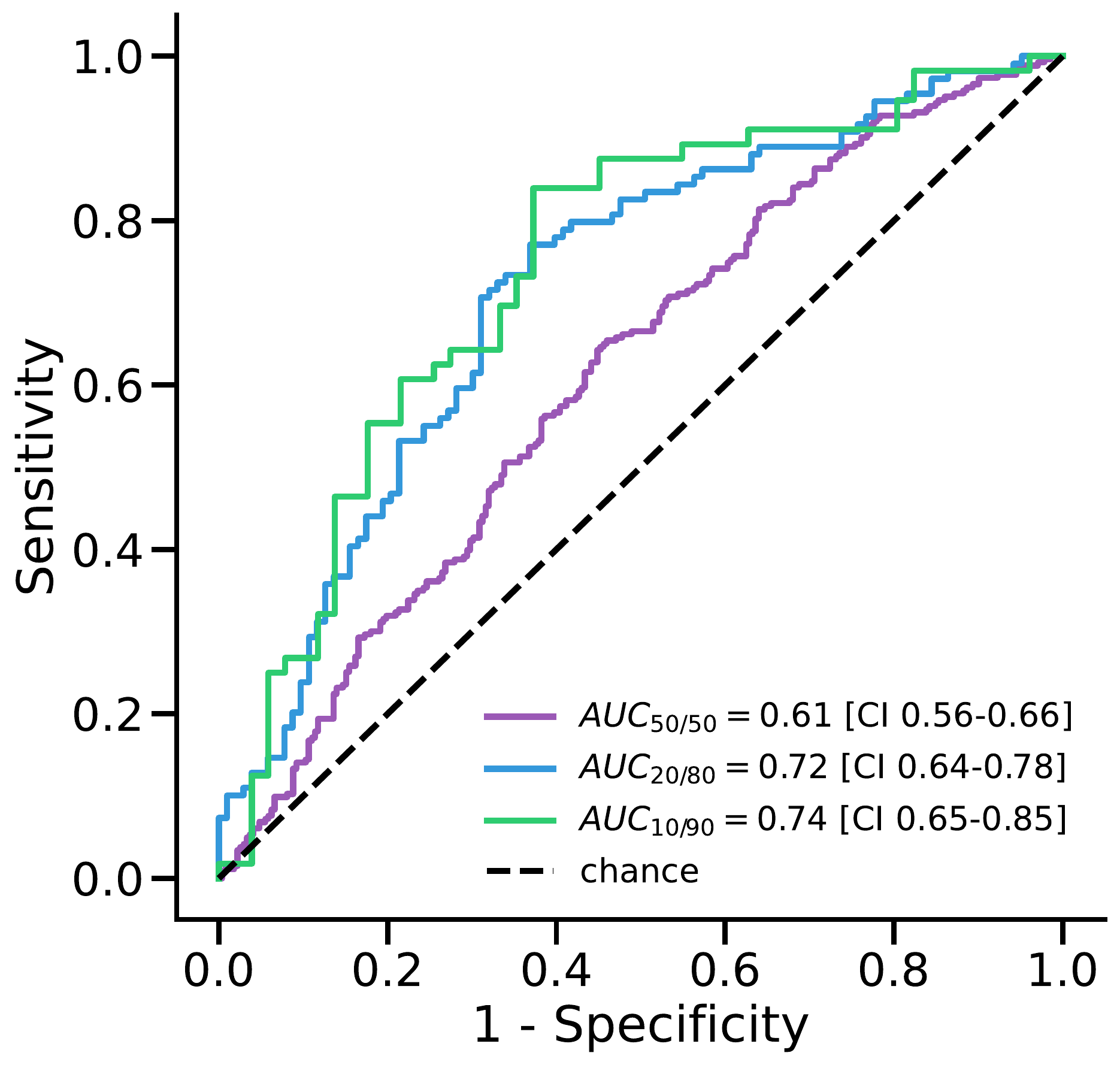}
        \caption{}
        \label{fig:FII_binary:b}
    \end{subfigure}
 
    \caption{\textbf{Evaluation in predicting the magnitude and direction of the VO2max change between the present and the future.} \textbf{(a)} Distribution of the $\Delta$ of VO2max in the present and the future. The shaded areas represent different binary bins that are used as outcomes, increasingly focusing on the extremes of this distribution. \textbf{(b)} ROC AUC performance in predicting the three $\Delta$ outcomes as shown on the left hand side.  Brackets represent 95\% CIs. } 
    \label{fig:FII_binary}
\end{figure*}

Motivated by the moderate predictability of the fine-grained delta of VO$_{2}max$, we formulated this problem as a classification task. A visual representation of this task can be found in Figure~\ref{fig:FII_binary:a}. By inspecting the distribution of the difference (delta) of current-future VO$_{2}max$ on the training set, we split it to 2 halves (50\% quantiles) \npj{of equally balanced data} and set these as prediction outcomes. The purpose of this task is to assess the \textit{direction} of individual change of fitness. We report an area under the curve (AUC) of 0.61 in predicting the direction of change ($N=2675$). We also \npj{investigated equal numbers of} participants on the tails of the change distribution which indicates participants who unterwent substantial and dramatic change in fitness over the period of time between Fenland I and Fenland II ($\approx$ 7 years). In this case, \npj{we picked participants from 80\%/20\% (substantial) and 90\%/10\% (dramatic) quantiles of the outcome distribution}. The results from these experiments show that the models can distinguish between substantial fitness change with an AUC of 0.72 ($N=1068$) and between dramatic fitness change with an AUC of 0.74 ($N=535$). All AUC curves can be found in Figure~\ref{fig:FII_binary:b}.

\begin{table}
\caption{\textbf{Evaluation of predicting fine-grained VO2max in the present and the future with the Fenland II repeats cohort using covariates of Fenland I}. Dense model results. (*the Delta outcome is in a different unit and hence a direct comparison with raw VO2max results might not apply).}

\resizebox{0.5\textwidth}{!}{
\begin{tabular}{@{}lllllcl@{}}
\toprule
\multicolumn{2}{c}{\multirow{2}{*}{Outcomes}}                      & \multicolumn{3}{c}{Evaluation Metrics {[}95\% CI{]}}                          & \multicolumn{1}{l}{N (train+val / test set)}                                      &  \\ \cmidrule(l){3-7} 
\multicolumn{2}{c}{}                                               & \multicolumn{1}{c}{R$^{2}$}  & \multicolumn{1}{c}{Corr} & \multicolumn{1}{c}{RMSE} & \multirow{7}{*}{\begin{tabular}[c]{@{}c@{}}2675\\  (2140/535)\end{tabular}} &  \\ \cmidrule(r){1-5}
\multicolumn{2}{l}{\textbf{Wearable Sensors + RHR + Anthro.}} &                         &                          &                          &                                                                             &  \\
                     & Current VO2max                              & 0.652 {[}0.606-0.695{]} & 0.815 {[}0.783-0.846{]}  & 2.959 {[}2.742-3.201{]}  &                                                                             &  \\
\multicolumn{2}{l}{\textbf{}}                                      &                         &                          &                          &                                                                             &  \\
                     & Future VO2max                               & 0.499 {[}0.431-0.55{]}  & 0.721 {[}0.67-0.759{]}   & 3.673 {[}3.421-3.916{]}  &                                                                             &  \\
\multicolumn{2}{l}{}                                               &                         &                          &                          &                                                                             &  \\
                     & Delta (Current - Future)*                   & 0.081 {[}0.02-0.078{]}  & 0.233 {[}0.159-0.307{]}  & 3.175 {[}2.923-3.41{]}   &                                                                             &  \\ \bottomrule
 \label{tab:fenII_results}
\end{tabular}
}
\end{table}

\subsection*{Enabling adaptive cardiorespiratory fitness inferences} 

\begin{figure*}
    \begin{subfigure}{.55\textwidth}
        \centering
        \includegraphics[scale=0.5]{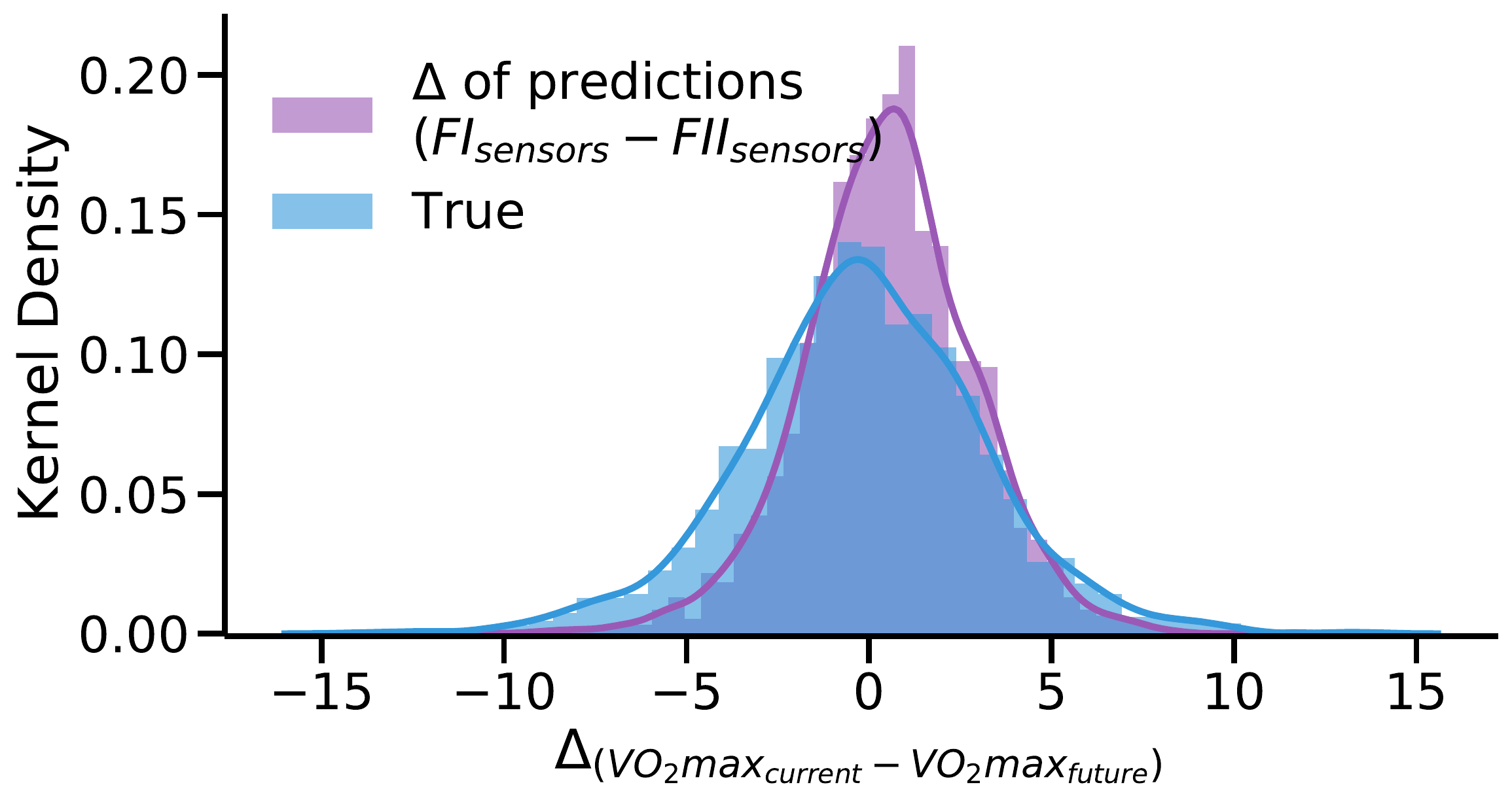}
        \caption{}
        \label{fig:adaptability:a}
    \end{subfigure}
    %
    %
    \begin{subfigure}{.45\textwidth}
        \centering
        \includegraphics[scale=0.4, angle=0]{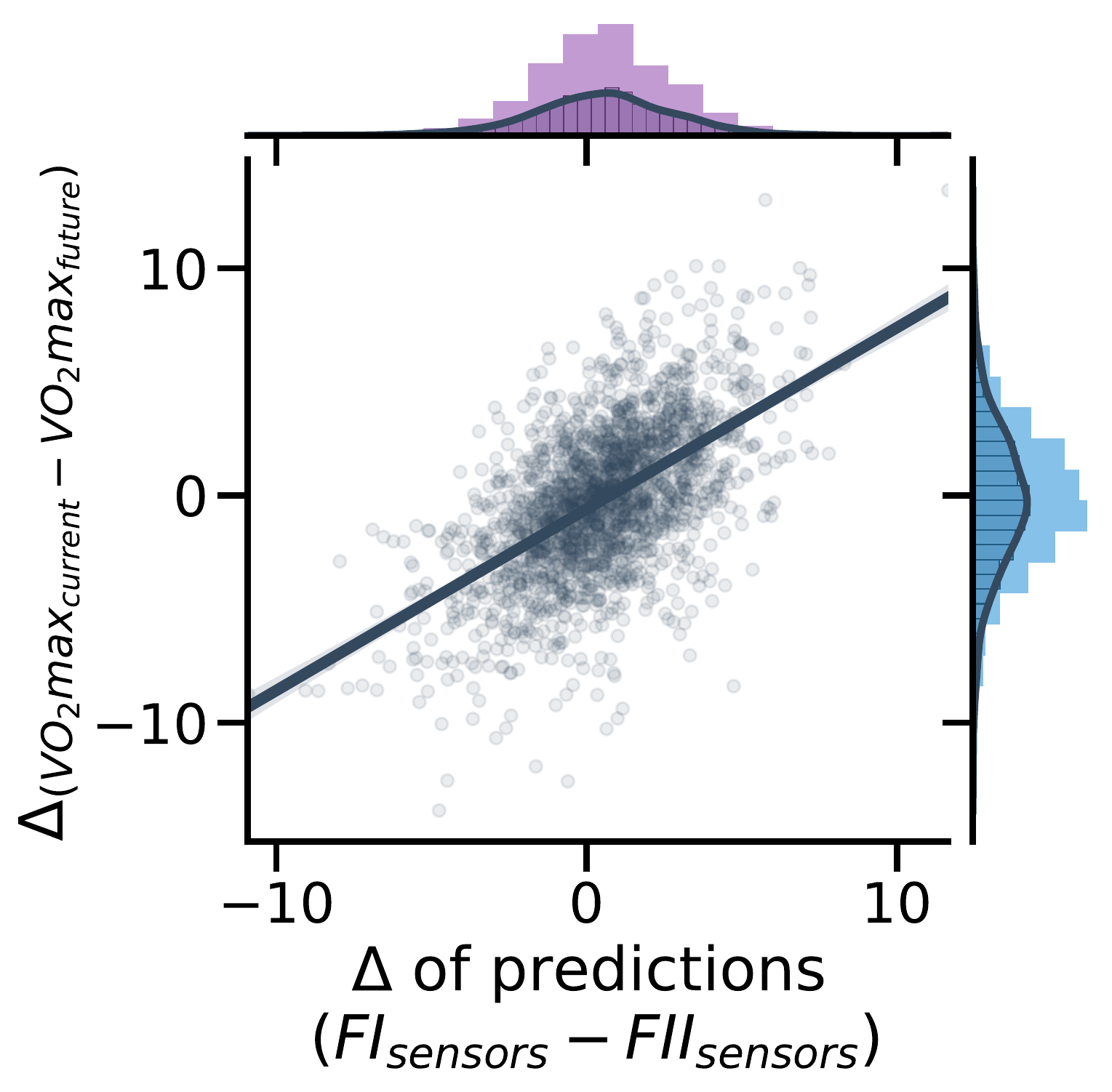}
        \caption{}
        \label{fig:adaptability:b}
    \end{subfigure}
    
    \vspace{5pt}
    
    \begin{subfigure}{.45\textwidth}
        \centering
        \includegraphics[scale=0.35, angle=0]{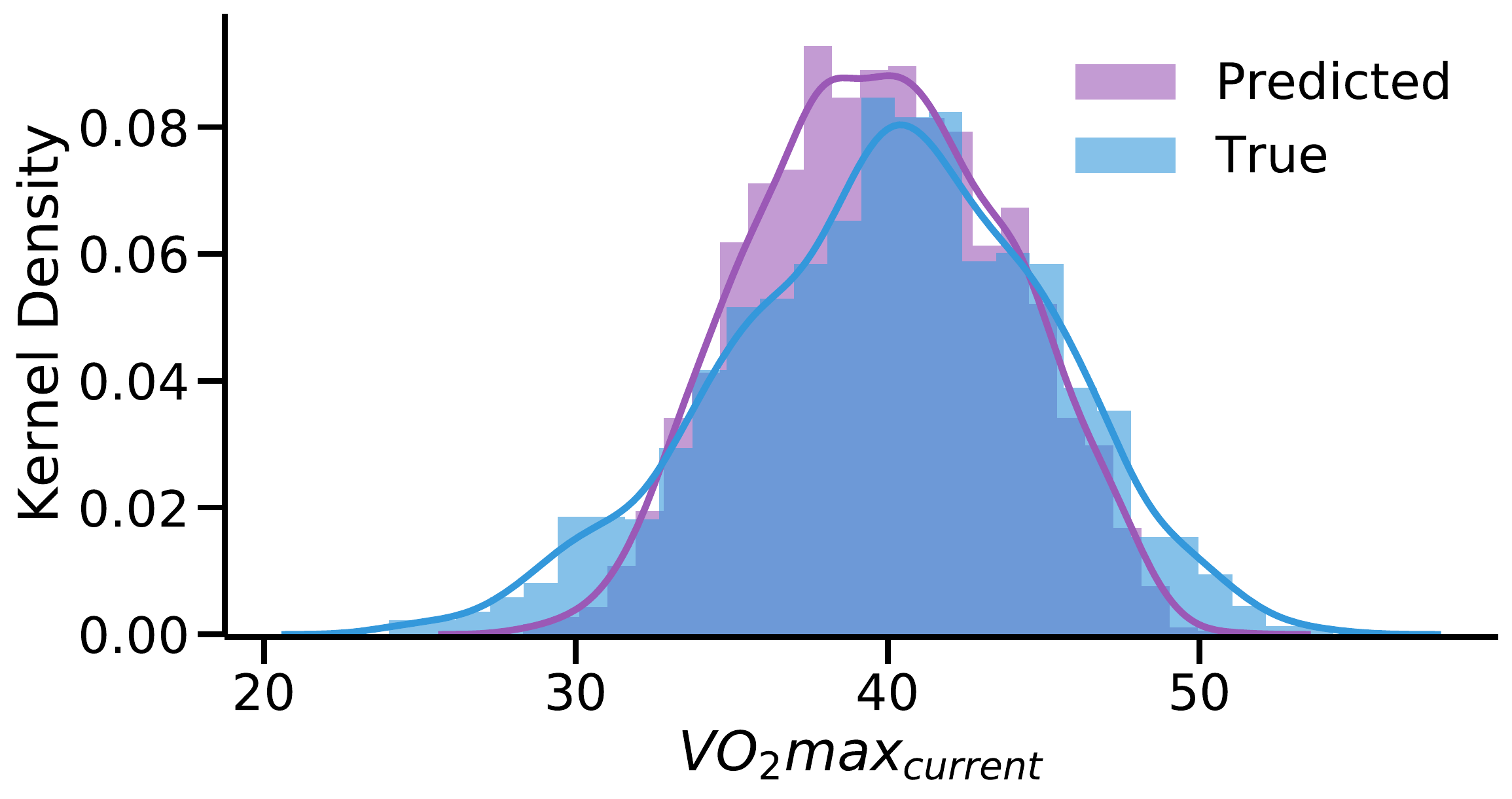}
        \caption{}
        \label{fig:adaptability:c}
    \end{subfigure}
    \begin{subfigure}{.45\textwidth}
        \centering
        \includegraphics[scale=0.35, angle=0]{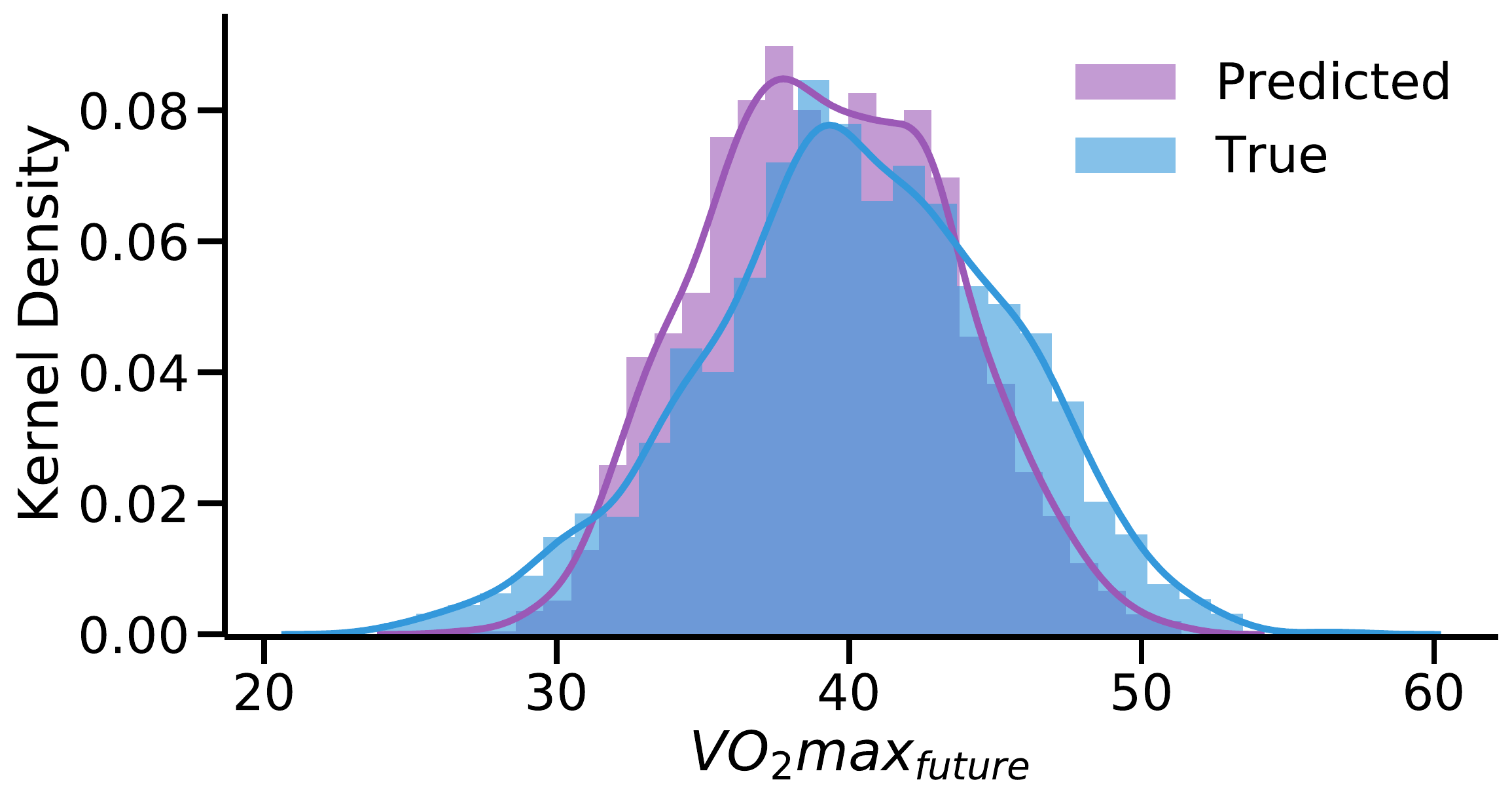}
        \caption{}
        \label{fig:adaptability:d}
    \end{subfigure}
 
    \caption{\textbf{Assessing model robustness over time using new sensor data from Fenland II repeats}. By matching the populations who provided sensor data for both cohorts (\textit{N}=2,042) we passed them through the trained model from Task 1. \textbf{(a-b)} Calculating the difference ($\Delta$) of the predictions juxtaposed with the true difference of fitness over the years with a correlation of $\Delta$ of predicted and true VO2max ($r=0.57$, $p<0.005$). \textbf{(c-d)} Comparison of predicted and true VO2max using FI and FII covariates (sensors, RHR, anthro.), respectively.
    } 
    \label{fig:adaptability}
\end{figure*}

\npj{For the final task, we assessed whether the trained models can pick up change using new sensor data from Fenland II, considering that obtaining new wearable data is relatively easy since these devices are becoming increasingly pervasive}. The intuition behind this task is to evaluate the generalizability of the models over time. We first matched the populations that provided sensor data for both cohorts ($N=2,042$) and applied the trained model from Task 1 in order to produce VO$_{2}$max inferences. We then compared the predictions with the respective ground truth (current and future VO$_{2}max$). The true and predictive distributions are shown in Figures~\ref{fig:adaptability:c} and~\ref{fig:adaptability:d}.  Through this procedure, we found that the model achieves an $r=0.84$ for VO$_{2}max$ future prediction and an $r=0.82$ for VO$_{2}max$ current prediction (validating our Task 1 results). In other words, if we have access to wearable sensor data and other information from the future time, we can reuse the already trained model from Fenland I to accurately infer fitness with minimal loss of accuracy over time, even though this is new sensor data from a completely separate (future) week. 

Last, we calculated the delta of the predictions and compared it to the actual delta of fitness over the years. This task showed that the models tend to focus mostly on positive change and under-predict when participants' fitness deteriorates over the years (Figures~\ref{fig:adaptability:a}, \ref{fig:adaptability:b}). The overall correlation between the delta of the predictions with the ground truth is significant (r=0.57, p<0.005).


\section*{Discussion}

Cardiorespiratory fitness declines with age independently of changes to body composition, and low cardiorespiratory fitness is associated with poor health outcomes~\cite{schmid2015cardiorespiratory,schuch2016lower,laukkanen2004predictive,mandsager2018association}. As such, having the capacity to predict whether CRF would decline in excess of natural aging could be valuable to clinicians when tailoring therapeutic interventions. Here we have developed a deep learning framework for predicting CRF and changes in CRF over time. Our framework estimates VO$_{2}$max by combining learned features from heart rate and accelerometer free-living data extracted from wearable sensors with anthropometric measures. To evaluate our framework's performance, VO$_{2}$max estimates were compared with VO$_{2}$max values derived from a submaximal exercise test \cite{gonzales2020estimating}.  Free-living and exercise test data were collected at a baseline investigation in 11,059 participants (Fenland I). A subset of those participants (n=2,675) completed another exercise test at a follow-up investigation approximately seven years later (Fenland II). This study design allowed us to address three questions: 1) Do baseline estimates of VO$_{2}$max from the deep learning framework agree with VO$_{2}$max values measured from exercise testing at baseline?, 2) Can the framework learn features from heart rate and accelerometer free-living data collected at baseline that predict VO$_{2}$max measured at follow-up?, and 3) Can the framework be used to predict the magnitude of change in VO$_{2}$max from baseline to follow-up? 

In the VO$_{2}$max estimation tasks, our model demonstrated strong agreement with VO$_{2}$max measured from the submaximal exercise test at baseline ($r$: 0.82) as well as for the longitudinal, follow-up visit ($r$: 0.72).  We were also able to distinguish between substantial and dramatic changes in CRF (AUCs 0.72 and 0.74, respectively). Finally, we further evaluated the initial model on new input data by feeding Fenland II free-living data along with updated heart rate and anthropometrics to the model, showing that it is able to adapt and monitor change over time. We evaluated the inference capabilities of the model in the difference (delta) between the current (Fenland I) and future (Fenland II)  VO$_{2}$max for those participants that came back approximately  7 years later. For this last task, the model produced outcomes that translated to a 0.57 correlation between the delta of predicted and delta of true VO$_{2}$max.


The application of our work to other cohort and longitudinal studies is of particular importance as serial measurement of cardiorespiratory fitness has significant prognostic value in clinical practice. Small increases in fitness are associated with reduced cardiovascular disease mortality risk 
and better clinical outcomes in patients with heart failure 
and type 2 diabetes \cite{jakicic2013four}. Nevertheless, routine measurement of fitness in clinical practice is rare due to the costs and risks of exercise testing. Non-exercise based regression models can be used to estimate changes in fitness in lieu of serial exercise testing. It is unclear, however, the extent to which changes in fitness detected with such models reflect true changes in exercise capacity. Here, we relied on the relationship between CRF and heart rate responses to different levels of physical activity at submaximal, real-life conditions captured through wearable sensors. Using deep learning techniques, we have developed a non-exercise based fitness estimation approach that can be used not only to accurately infer current VO$_{2}$max, but also to do so in the settings of a future cohort, where the model did not require any retraining, just influx of new data. Further, we show that the model can also be used to infer the changes in CRF that occurred during the $\approx$ 7 year time span between Fenland I and II. 

\begin{table*}
\caption{\npj{\textbf{Sensitivity analysis of the comprehensive Dense model with regards to anthropometrics}. Breakdown of sex, age, weight, BMI, and height based on either the median value of the testing set or the available categories. Bold font denotes best values (low for Error metrics, high for R2 and Corr). Note: the groups are not balanced because the median value(s) were assigned to the first ($\leq$) group.}}
\resizebox{0.99\textwidth}{!}{
\begin{tabular}{lllllllll}
\hline
\multicolumn{2}{c}{\multirow{2}{*}{Features}} & \multicolumn{7}{c}{Evaluation Metrics {[}95\% CI{]}}                                                                                                                                                                                               \\ \cline{3-9} 
\multicolumn{2}{c}{}                          & \multicolumn{1}{c}{R2}           & \multicolumn{1}{c}{Corr}         & \multicolumn{1}{c}{RMSE}         & MSE                              & MAE                              & STD of MAE                       & MAPE                             \\ \hline
\multicolumn{2}{l}{\textbf{Sex}}              &                                  &                                  &                                  &                                  &                                  &                                  &                                  \\
           & Male (N=1303)                    & 0.592 {[}0.553-0.627{]}          & 0.772 {[}0.745-0.797{]}          & 2.826 {[}2.671-2.986{]}          & 7.987 {[}7.133-8.917{]}          & \textbf{2.153 {[}2.059-2.259{]}} & 1.831 {[}1.687-1.975{]}          & \textbf{0.053 {[}0.050-0.056{]}} \\
           & Female (N=1372)                  & 0.599 {[}0.567-0.629{]}          & 0.779 {[}0.756-0.799{]}          & 2.974 {[}2.855-3.096{]}          & 8.845 {[}8.148-9.584{]}          & 2.350 {[}2.259-2.444{]}          & 1.822 {[}1.727-1.936{]}          & 0.065 {[}0.062-0.068{]}          \\
\multicolumn{2}{l}{\textbf{Age}}              &                                  &                                  &                                  &                                  &                                  &                                  &                                  \\
           & $\leq$ 48 years (N=1405)              & 0.680 {[}0.650-0.708{]}          & 0.829 {[}0.809-0.847{]}          & 2.803 {[}2.652-2.950{]}          & \textbf{7.855 {[}7.033-8.702{]}} & 2.161 {[}2.064-2.259{]}          & 1.785 {[}1.646-1.912{]}          & 0.055 {[}0.052-0.058{]}          \\
           & $\textgreater$ 48 years (N=1270)   & 0.607 {[}0.566-0.642{]}          & 0.780 {[}0.753-0.804{]}          & 3.010 {[}2.862-3.143{]}          & 9.059 {[}8.190-9.879{]}          & 2.357 {[}2.253-2.450{]}          & 1.871 {[}1.744-1.983{]}          & 0.064 {[}0.061-0.066{]}          \\
\multicolumn{2}{l}{\textbf{Weight}}           &                                  &                                  &                                  &                                  &                                  &                                  &                                  \\
           & $\leq$ 75.3 kg (N= 1340)              & 0.639 {[}0.604-0.669{]}          & 0.803 {[}0.781-0.824{]}          & 2.982 {[}2.838-3.139{]}          & 8.891 {[}8.056-9.852{]}          & 2.322 {[}2.222-2.418{]}          & 1.870 {[}1.737-2.021{]}          & 0.061 {[}0.058-0.064{]}          \\
           & $\textgreater$ 75.3 kg (N=1335)    & \textbf{0.698 {[}0.668-0.724{]}} & \textbf{0.837 {[}0.819-0.854{]}} & \textbf{2.822 {[}2.684-2.948{]}} & 7.961 {[}7.205-8.692{]}          & 2.185 {[}2.085-2.276{]}          & 1.785 {[}1.668-1.900{]}          & 0.057 {[}0.054-0.060{]}          \\
\multicolumn{2}{l}{\textbf{BMI}}              &                                  &                                  &                                  &                                  &                                  &                                  &                                  \\
           & $\leq$ 25.5 kg/m2 (N= 1338)           & 0.640 {[}0.604-0.670{]}          & 0.804 {[}0.779-0.825{]}          & 2.973 {[}2.814-3.146{]}          & 8.838 {[}7.917-9.899{]}          & 2.303 {[}2.208-2.414{]}          & 1.880 {[}1.733-2.044{]}          & 0.059 {[}0.056-0.062{]}          \\
\textbf{}  & $\textgreater$ 25.5 kg/m2 (N=1337) & 0.683 {[}0.653-0.711{]}          & 0.828 {[}0.809-0.845{]}          & 2.831 {[}2.706-2.958{]}          & 8.016 {[}7.323-8.748{]}          & 2.205 {[}2.110-2.300{]}          & \textbf{1.775 {[}1.674-1.879{]}} & 0.059 {[}0.056-0.062{]}          \\
\multicolumn{2}{l}{\textbf{Height}}           &                                  &                                  &                                  &                                  &                                  &                                  &                                  \\
           & $\leq$ 1.70m (N=1343)                 & 0.622 {[}0.588-0.651{]}          & 0.792 {[}0.769-0.811{]}          & 2.962 {[}2.831-3.092{]}          & 8.771 {[}8.014-9.563{]}          & 2.315 {[}2.213-2.405{]}          & 1.847 {[}1.743-1.968{]}          & 0.064 {[}0.061-0.066{]}          \\
\textbf{}  & $\textgreater$ 1.70m (N=1332)      & 0.619 {[}0.582-0.651{]}          & 0.790 {[}0.767-0.812{]}          & 2.843 {[}2.714-2.996{]}          & 8.080 {[}7.368-8.973{]}          & 2.193 {[}2.107-2.298{]}          & 1.809 {[}1.688-1.952{]}          & 0.054 {[}0.052-0.057{]}          \\ \hline
\label{tab:breakdown}
\end{tabular}
}
\end{table*}

Our proposed deep learning approach outperforms traditional non-exercise models, which are the state-of-the art in the field and rely on simple variables inputted to a linear model. Importantly, our model is able to take week-level information from each participant and combine it with various anthropometrics and bio-markers such as the RHR, providing a truly personalized approach for CRF inference generation. The approach we present here outperforms traditional non-exercise models, which are considered state-of-the-art methods for longitudinal monitoring  and highlights the potential of wearable sensing technologies for digital health monitoring. 
An additional application of our work is the potential routine estimation of VO$_{2}$max in clinical settings, given the strong association between estimated CRF levels and CVD health outcomes \cite{qui2021crfmeta}.

This study has several limitations worthy of recognition. First, the validity of the deep learning framework was assessed by comparing estimated VO$_{2}$max values with VO$_{2}$max values derived from a submaximal exercise test. Ideally, one would use VO$_{2}$max values directly measured during a maximal exercise test to establish the ground truth for cardiorespiratory fitness comparisons.  Maximal exercise tests, however, are problematic when used in large population-based studies because they may be unsafe for some participants and, consequently, induce selection bias. The submaximal exercise test used in the Fenland Study was well tolerated by study participants and demonstrated acceptable validity against direct VO2max measurements \cite{gonzales2020estimating}. \revised{Submaximal tests are also utilized to validate popular wearable devices such as the Apple Watch \cite{apple}.} We are therefore confident that VO$_{2}$max values estimated from the deep learning framework reflect true cardiorespiratory fitness levels.

\revised{To investigate this limitation, we validate our models with 181 participants from the UK Biobank Validation Study (BBVS) who were recruited from the Fenland study. These participants completed an independent \textit{maximal} exercise test, where VO$_{2}$max was directly measured. Taking into account that the BBVS cohort is less fit (VO$_{2}$max=32.9$\pm$7) compared to the cohort the model was trained on (Fenland I, VO$_{2}$max=39.5$\pm$5), we observe that the model over-predicts with a mean prediction of VO$_{2}$max=39.9, RMSE=8.998. This is expected because the range of VO$_{2}$max seen during training did not include participants with VO$_{2}$max below 25. Even when looking at women in isolation -who perform lower than men in these tests-, they had a  VO$_{2}$max=37.4$\pm$4.7 in Fenland I (see Figure \ref{fig:flowchart}), which is still significantly higher than the average participant of BBVS. This is a common distribution/label shift issue where the model encounters an outcome which is out of its training data range and is still an open problem in statistical modelling \cite{lipton2018detecting}.
To partially mitigate this issue, we match BBVS's fitness to have similar statistics to the training set of original model (Fenland I: VO$_{2}$max=39$\pm$5, matched BBVS: VO$_{2}$max=39$\pm$4) and observe an RMSE=5.19 (see Figure \ref{fig:BVS}). This result is still not on par with our main results in Fenland but shows the impact of including very low-fitness participants in this validation study. To improve future CRF models, we believe that population-scale studies should focus on including low-fitness participants along with the general population. }

Putting our results in context, we are confident that the level of accuracy of our methods is acceptable for use in population scale or even commercial wearables. For example, in Table \ref{tab:breakdown}, we observe that the MAE of all subgroups is between 2.1-2.3. Future directions to improve these results include transfer learning and domain adaptation, particularly tailored to the problem of distribution shift, as discussed in the previous paragraph. Comparing to smaller studies such as the Apple Watch study which was conducted in more constrained environments that required users to log their workouts \cite{apple}, we see that the reported MAE is 1.4. We believe this difference is reasonable due to the completely free-living data we incorporate, bringing our study's evaluation setup closer to the real-world.

In this paper, we developed deep learning models utilising wearable data and other bio-markers to predict the gold standard of fitness (VO$_{2}max$) and achieved strong performance compared to other traditional approaches. Cardio-respiratory fitness is a well-established predictor of metabolic disease and mortality and our premise is that modern wearables capture non-standardised dynamic data which could improve fitness prediction. Our findings on a population of 11,059 participants showed that the combination of all modalities reached an $r$ = 0.82, when compared to the ground truth in a holdout sample. Additionally, we show the adaptability and applicability of this approach for detecting fitness change over time in a longitudinal subsample (\textit{n = 2,675}) who repeated measurements after 7 years. Last, the latent representations that arise from this model pave the way for fitness-aware monitoring and interventions at scale. It is often said that \textit{"If you cannot measure it, you cannot improve it"}. Cardio-fitness is such an important health marker, but until now we did not have the means to measure it at scale. Our findings could have significant implications for population health policies, finally moving beyond weaker health proxies such as the BMI.

\section*{Methods}
\subsection*{Study description}

The Fenland study is a population-based cohort study designed to investigate the independent and interacting effects of environmental, lifestyle and genetic influences on the development of obesity, type 2 diabetes and related metabolic disorders. Exclusion criteria included prevalent diabetes, pregnancy or lactation, inability to walk unaided, psychosis or terminal illness (life expectancy of $\leq$ 1 year at the time of recruitment) .

The Fenland study has two distinct phases. Phase I, during which baseline data was collected from 12,435 participants, took place between 2005 and 2015. Phase II was launched in 2014 and involved repeating the measurements collected during Phase I, alongside the collection of new measures. All participants who had consented to being re-contacted after their Phase I assessment were invited to participate in Phase II. At least 4 years must have elapsed between visits. As a result of this stipulation, recruitment to Phase II is ongoing. A flowchart of the analytical sample by each one of the study tasks is provided in Figure~\ref{fig:flowchart}.

\begin{figure}
    \centering
    \includegraphics[width=0.99\linewidth]{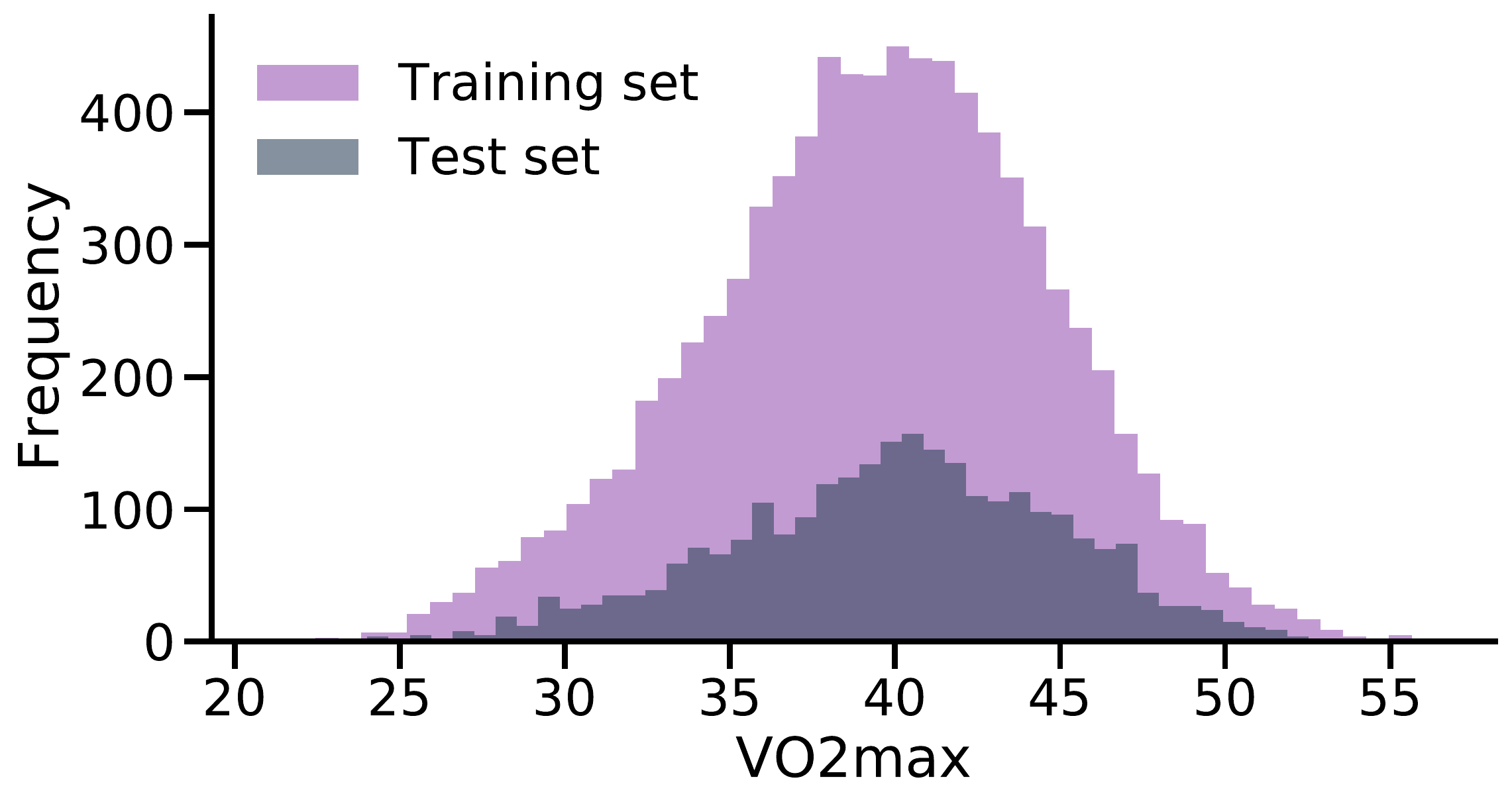}
	\caption{\textbf{Distribution of VO2max in the training and test sets in Fenland I cohort.} Both sets display similar ranges of values, making sure that inferences based on the test set are robust. This plot refers to Task's 1 train and test sets.}
    \label{fig:train_test_llabel} 
\end{figure}

After a baseline clinic visit, participants were asked to wear a combined heart rate and movement chest sensor Actiheart, CamNtech, Cambridgeshire, UK) for 6 complete days. For this study, data from 11,059 participants was included after excluding participants with insufficient or corrupt data or missing covariates as shown in Figure~\ref{fig:flowchart}. A subset of 2,675 of the study participants returned for the second phase of the study,  after a median (interquartile range) of 7 (5-8) years, and underwent a similar set of tests and protocols, including wearing the combined heart rate and movement sensing for 6 days. All participants provided written informed consent and the study was approved by the University of Cambridge, NRES Committee - East of England Cambridge Central committee. All experiments and data collected were done in accordance with the declaration of Helsinki.

\subsection*{Study procedure}

Participants wore the Actiheart \textit{wearable ECG} which measured heart rate and  movement recording at 60-second intervals~\cite{brage2005reliability}. The Actiheart device was attached to the chest at the base of the sternum by two standard ECG electrodes. Participants were told to wear the monitor continuously for 6 complete days and were advised that these were waterproof and could be worn during showering, sleeping or exercising. During a lab visit, all participants performed a treadmill test that was used to establish their individual response to a \textit{submaximal test}, informing their $VO_{2}max$ (maximum rate of oxygen consumption measured during incremental exercise)~\cite{brage2007hierarchy}. RHR was measured with the participant in a supine position using the Actiheart device. HR was recorded for 15 minutes and RHR was calculated as the mean heart rate measured during the last 3 minutes. Our RHR is a combination of the Sleeping HR measured by the ECG over the free-living phase and the RHR as described above.  

\subsection*{Cardiorespiratory fitness assessment}
VO$_{2}max$ was predicted in study participants using a previously validated submaximal treadmill test ~\cite{gonzales2020estimating}. Participants exercised while treadmill grade and speed were progressively increased across several stages of level walking, inclined walking, and level running. The test was terminated if one of the following criteria were met: 1) the participant wanted to stop, 2) the participant reached 90$\%$ of age-predicted maximal heart rate (208-0.7*age) ~\cite{tanaka2001age}, 3) the participants exercised at or above 80$\%$ of age-predicted maximal heart rate for 2 minutes. \revised{Details about the fitness characteristics of the cohort and the validation of the submaximal test are provided elsewhere ~\cite{gonzales2022descriptive}}.  

\revised{To further validate the models trained on submaximal VO$_{2}max$, we employ the external cohort UK Biobank Validation Study (BBVS)~\cite{gonzales2022descriptive}. We recruited 105 female (mean age: 54.3y$\pm$7.3) and 86 male (mean age: 55.0y$\pm$6.5) validation study participants and VO$_{2}max$ was directly measured during an independent maximal exercise test, which was completed to exhaustion. Some maximal exercise test data were excluded because certain direct measurements were anomalous due to testing conditions (N=10). BBVS participants completed the same free-living protocol as in Fenland and we collected similar sensor and antrhopometrics data which were processed with the same way as in Fenland (see next section).
}

\subsection*{Free-living wearable sensor data processing}

Participants were excluded from this analysis if they had less than 72 hours of concurrent wear data (three full days of recording) or insufficient individual calibration data (treadmill test-based data). All heart rate data collected during free-living conditions underwent pre-processing for noise filtering.
Non-wear detection procedures were applied and any of those non-wear periods were excluded from the analyses.
This algorithm detected extended periods of non-physiological heart rate concomitantly with extended ($>$ 90 minutes) periods that also registered no movement through the device's accelerometer. We converted movement these intensities into standard metabolic equivalent units (METs), through the conversion 1 MET = 71 J/min/kg (~3.5 ml O$_{2} \cdot$ min$^{-1} \cdot$ kg$^{-1}$). These conversions where then used to determine intensity levels with $\leq 1.5$METs classified as sedentary behaviour, activities between 3 and 6 METs were classified as moderate to vigorous physical activity (MVPA) and those $>6$METs were classified as vigorous physical activity (VPA). 

Since the season can have a big impact on physical activity considering on how it affects workouts, sleeping patterns, and commuting patterns, we encoded the sensor timestamps using \textit{cyclical temporal features} $T_f$.
Here we encoded the month of the year as $(x,y)$ coordinates on a circle:

\vspace{-.1cm}
\noindent\begin{minipage}{.50\linewidth}
\begin{equation}
    T_{f_1} = sin \Big(  \frac{2*\pi*t}{max(t)} \Big)
\end{equation}
\end{minipage}%
\noindent\begin{minipage}{.50\linewidth}
\begin{equation}
    T_{f_2} = cos \Big(\frac{2*\pi*t}{max(t)} \Big)
\end{equation}
\end{minipage}%

where $t$ is the relevant temporal feature (month). The intuition behind this encoding is that the model will "see" that e.g. December (12th) and January (1st) are 1 month apart (not 11). Considering that the month might change over the course of the week, we use the month of the first time-step only.
Additionally, we extracted summary statistics from the following sensor time-series: raw acceleration, HR, HRV, Aceleration-derived Euclidean Norm Minus One, and Acceleration-derived Metabolic Equivalents of Task. Then, for every time-series we extracted the following variables which cover a diverse set of attributes of their distributions: mean, minimum, maximum, standard deviation, percentiles (25\%, 50\%, 75\%), and the slope of a linear regression fit. The rest of variables (anthropometrics and RHR) are used as a single measurement.

In total, we derived a comprehensive set of 68 features using the Python libraries Pandas 
and Numpy.
A detailed view of the variables is provided in Table \ref{tab:variables}.

\begin{figure*}
    \centering
    \includegraphics[width=0.8\linewidth]{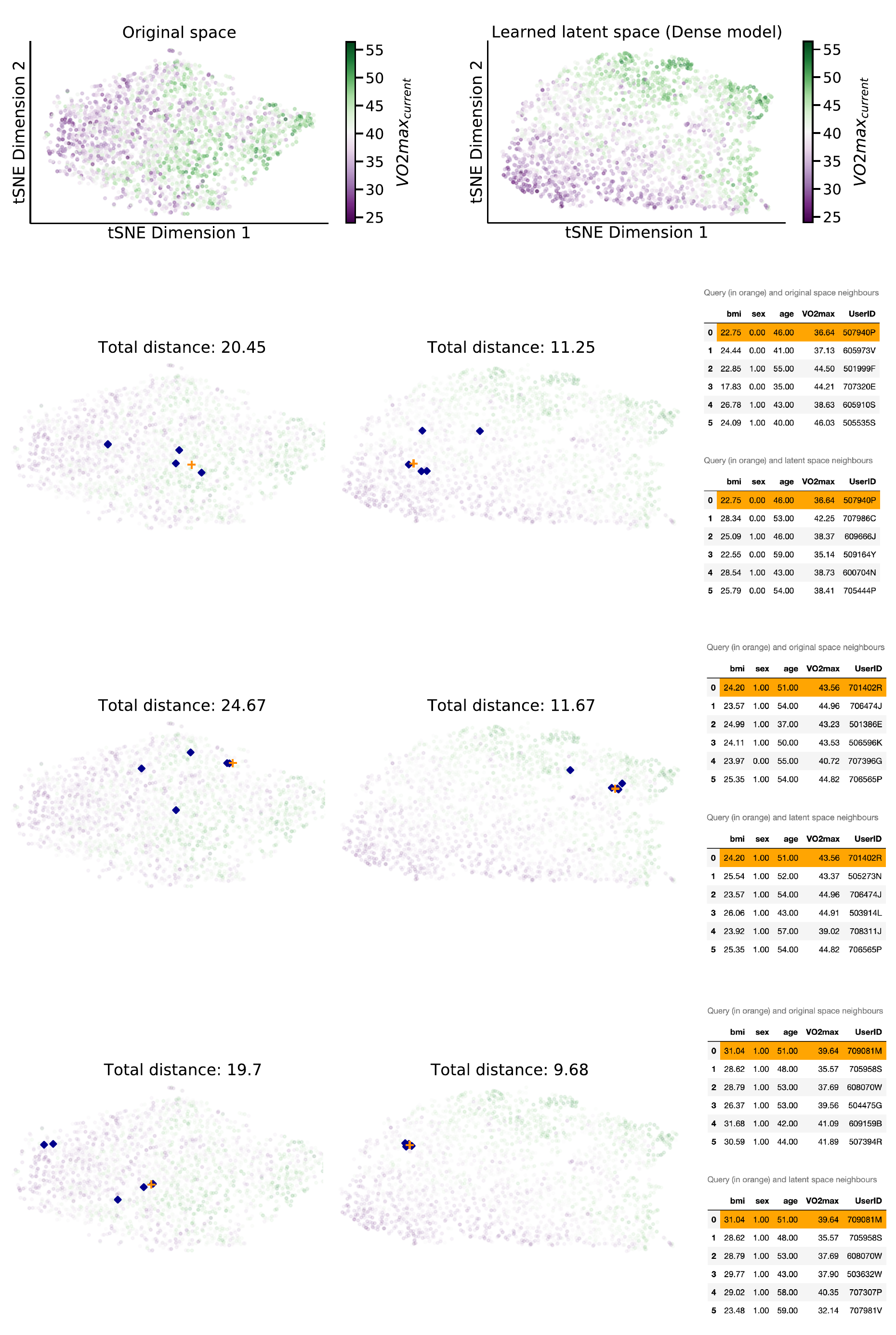}
	\caption{\npj{\textbf{Fitness subtyping through latent neighbours.} tSNE projection of the original feature vector (Fenland I testing set, \textit{Sensors + RHR + Anthro}.) and the latent space of the Dense model after training. \textbf{Top}: The original data presents some clusters but the outcome is not clearly linearly separable.  The model activations on the penultimate layer of the neural network capture the continuum of low-high VO$_{2}max$ both locally and globally. \textbf{Bottom}: Starting from a query participant (+) we retrieve the 5 nearest neighbours in the original and latent space and list their details on the tables on the right. The total distance of each query to each neighbour is listed in each subplot. Transparency has been applied to combat crowding and the colorbar is centered on the median value to illustrate extreme cases. The VO$_{2}max$ label is used only for color-coding purposes (the projection is label-agnostic). Every participant is a dot.  
    }}
    \label{fig:tsne} 
\end{figure*}



\subsection*{Deep learning models}

We developed deep neural network models that are able to capture non-linear relationships between the input data and the respective outcomes. Considering the high-sampling rate of the sensors (1 sample/minute) after aligning HR and Acceleration modalities, it is impossible to learn patterns with such long dependencies (a week of sensor data includes more than 10,000 timesteps). Even the most well-tuned  recurrent neural networks cannot cope with such sequences and given the size of the training set (7,545 samples), the best option was to extract statistical features from the sensors and represent every participant-week as a row in a feature vector (see Fig. \ref{fig:study_hero}). This feature vector was fed to fully connected neural network layers which were trained with backpropagation. All deep learning models are implemented in Python using Tensorflow/Keras.

\noindent\textbf{Data preparation}. For Task 1 (see Figure \ref{fig:flowchart}), we matched the sensor data with the participants who had eligible lab tests. Then we split into disjoint train and test sets, making sure that participants from Fenland I go to the train set, while those from Fenland II go to the test set (see Figure \ref{fig:train_test_llabel}). This would allow to re-use the trained model from Task 1, with different sensor data from Fenland II participants. Intuitively, we train a model on the big population, and we evaluate it with two snapshots of another longitudinal population over time (Task 1 \& 3). After splitting, we normalize the training data by applying standard scaling (removing the mean and scaling to unit variance) and then  denoise it by applying Principal Components Analysis (PCA), retaining the components that explain 99.99\% of the variance. In practice, the original 68 features are reduced to 48. We save the \textit{fitted} PCA projection and scaler and we apply them individually to the test-set, to avoid information leakage across the sets. The same projection and scaler are applied to all downstream models (Task 2 and 3) to leverage the knowledge of the big cohort (Fenland I).

\noindent\textbf{Model architecture and training}. The main neural network (used in Task 1) receives a 2D vector of [users, features] and predicts a real value. For this work, we assume $\mathcal{\textit{{N}}}$ users and $\mathcal{\textit{{F}}}$ features of an input vector $\mathcal{\textbf{X}}$ = (${\textbf{x}_{1}}$,...,${\textbf{x}_{N}}$) $\in \mathbb{R}^{N \times F}$ and a target $VO_2max$ $\mathcal{\textbf{y}}$ = (${\textbf{y}_1}$,...,${\textbf{y}_{N}}$) $\in \mathbb{R}^{N}$. The network consists of two densely-connected feed forward layers with 128 units each. A dense layer works as follows: $output$ = $activation$ ($input$ $\cdot$ $kernel$ + $bias$), where activation is the element-wise activation function (the exponential linear unit in our case)
, kernel is a learned weights matrix with a Glorot uniform initialization, 
and bias is a learned bias vector. Each layer is followed by a \textit{batch normalization} 
operation, which maintains the mean output close to 0 and the output standard deviation close to 1. Also, dropout of 0.3 probability is applied to every layer, which randomly sets input units to 0 and helps prevent overfitting. 
Last, the final layer is a single-unit dense layer and the network is trained with the Adam optimizer 
to minimize the Mean Squared Error (MSE) loss, which is appropriate for continuous outcomes. We use a random 10\% subset of the train-set as a validation set. To combat overfitting, we train for 300 epochs with a batch size of 32 and we perform early stopping when the validation loss stops improving after 15 epochs and the learning rate is reduced by 0.1 every 5 epochs. All hyperparameters (\# layers, \# units, dropout rates, batch size, activations, and early stopping) were found after tuning on the validation set.

\noindent\textbf{Model differences across tasks}. Task 1 trains the main neural network of our study (see previous subsection). Task 2 re-trains an identical model to predict $VO_2max$  in the future (and the delta present-future). \npj{We note that the delta prediction task cannot be comparable with the models predicting the present and future outcomes. Essentially, the delta model predicts the difference between these two timepoints, which results in a range of values roughly from -10 to +10. This distribution is not normally distributed (Shapiro-Wilk test=0.991, p=0.002) and hence both linear and neural models cannot approximate the tails, with most of their predictions lying between -3 and +3. The negative/positive signs of this outcome make the error metrics not very interpretable. We don't believe this performance is caused by overfitting because the results of both linear and neural models are similar. This result motivated us to study the delta distribution as a binary problem.} When we re-frame this problem as a classification task (see Figure \ref{fig:FII_binary}), we use significantly fewer participants when we focus on the tails of the change distribution. Therefore, to combat overfitting, we train a smaller network with one Dense layer of 128 units and a sigmoid output unit, which is appropriate for binary problems. Instead of optimizing the MSE, we now minimize the binary cross-entropy. In all other cases --such as in Task 3 or when visualizing the latent space--, we do not train new models; the model which was trained in Task 1 is used in inference mode (prediction).

\subsection*{Prediction equations}
For reference, we compare our models results to traditional non-model equations, which rely on Body Mass, RHR, and Age. We incorporate the popular equation proposed by Uth and colleagues \cite{uth2004estimation}, which corresponds to $VO2max$ = 15.0 ($ml*minˆ-1$) * Body Mass (kg) *  (HRmax / HRrest), in combination with Tanaka's equation \cite{tanaka2001age} where $HRmax=208-0.7*age$. Other approaches rely on measurements such as the waist circumference, which however was not recorded in our cohorts.

\npj{\subsection*{Linear model} We begin our investigation by establishing a strong baseline with a linear regression model (as seen in Table \ref{tab:fenI_results}). We compare differenet combinations of input data and finally compare the comprehensive model with the Dense neural network. We use the Python sklearn implementation for the linear regression. }

\subsection*{Evaluation}

To evaluate the performance of the deep learning models which predict continuous values, we computed the root mean squared error (RMSE) \begin{math}
    = \sqrt{\frac{1}{\left | N_{test} \right |} \sum_{y  \in \mathcal{D}_{test}}  \sum^{N}_{t=1} (y_{t} - \hat{y}_{t})^{2}}
\end{math}, the coefficient of determination ($R^{2}$) \begin{math}
    = 1 - \frac{\sum^{N}_{t=1} (y_{t} - \hat{y}_{t})^{2}}{\sum^{N}_{t=1} (y_{t} - \bar{y}_{t})^{2}}
\end{math}, and the Pearson correlation coefficient \npj{for the majority of the analyses as they capture different properties of the error distributions in regression tasks. For the subgroup sensitivity analysis, we additionally employed the Mean Squared Error (MSE), Mean Absolute Error (MAE) and its standard deviation (STD of MAE), and the Mean Absolute Percentage Error (MAPE).} In most regression metrics, $y$  and $\hat{y}$ are the measured and predicted VO$_{2}max$ and $\bar{y}$ is the mean. For the binary models, we used the Area under the Receiver Operator Characteristic (AUROC or AUC) which evaluates the probability of a randomly selected positive sample to be ranked higher than a randomly selected negative sample.

\subsection*{Visualizing the latent space} The activations of the trained model allow us to understand the inner workings of the network and explore its latent space. We first pass the test-set of Task 1 through the trained model and retrieve the activations of the penultimate layer \cite{yosinski2015understanding}.
This is a 2D vector of [2675, 128] size, considering that the layer size is 128 and the participants of the test-set are 2675. Intuitively, every participant corresponds to a 128-dimensional point. In order to visualize this embedding, we apply tSNE \cite{van2008visualizing},
an algorithm for dimensionality reduction. For its optimization, we use a perplexity of 50, as it was suggested recently \cite{wattenberg2016use}. \npj{We calculated the k-nearest neighbours on both the original and latent spaces (using k=5) in a case study presented in Figure \ref{fig:tsne}. We also calculated the total euclidean distance of each query participant across all their neighbours, as a means of quantifying the proximity in the high-dimensional space.}

\subsection*{Statistical analyses}

We performed a number of sensitivity analyses to investigate potential sources of bias in our results. Full results of these sensitivity analyses are shown in the main text and corresponding Tables. In particular, we use bootstraping with replacement \npj{(500 samples)} to calculate 95\% Confidence Intervals when we report the performance of the models in the hold-out sets. 
Wherever we report p-values, we use the recently suggested threshold of $p<0.005$ for human studies \cite{benjamin2018redefine}.






\section*{Data availability}

All Fenland data used in our analyses is available from the MRC Epidemiology Unit at the University of Cambridge upon reasonable request (\url{http://www.mrc-epid.cam.ac.uk/research/studies/fenland/}).

\section*{Code availability}

Source code, data samples, and pre-trained models of this work are available on Github (\url{https://github.com/sdimi/cardiofitness/}).

\section*{Acknowledgements}

DS was supported by the Embiricos Trust Scholarship of Jesus College
Cambridge and the Engineering and Physical Sciences Research
Council through grant DTP (EP/N509620/1). IP-P was supported by GlaxoSmithKline and Engineering and Physical
Sciences Research Council through an iCase fellowship (17100053).

\section*{Author contributions statement}

DS, IP, SB, NW, CM designed the study. SB, NW collected the data on behalf of the Fenland Study. DS, IP are co-first authors and sourced, selected, and pre-processed the data. DS developed the models and produced the figures and tables. DS, IP performed the statistical analysis. 
IP, DS wrote the first draft of the manuscript.
All authors (DS, IP, TG, YW, SB, NW, CM) critically reviewed, contributed to the preparation of the manuscript, and approved the final version. All authors vouch for the data, analyses, and interpretations.

\section*{Competing interests}
The authors declare that they have no
competing financial interests.

\bibliography{VO2wearable}


\onecolumn
\renewcommand\thefigure{Suppl. \arabic{figure}}    
\renewcommand\thetable{Suppl. \arabic{figure}}    
\setcounter{figure}{0}    
\setcounter{table}{0}    
\section*{Supplementary information}

\begin{figure}[h]
    \begin{subfigure}{.5\textwidth}
        \centering
        \includegraphics[scale=0.35]{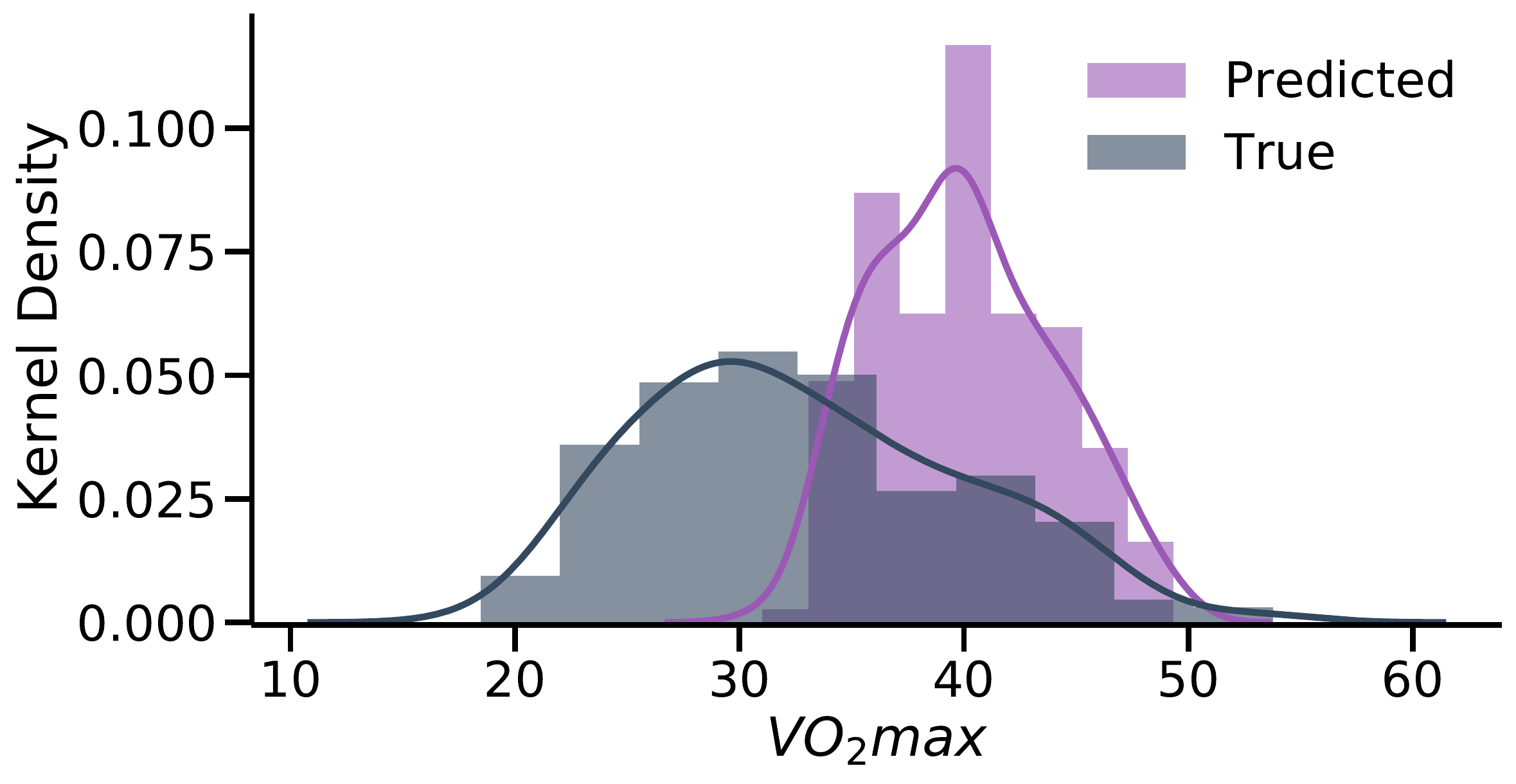}
        \caption{}
        \label{fig:BVS:a}
    \end{subfigure}
   \hspace{10pt}
    \begin{subfigure}{.5\textwidth}
        \centering
        \includegraphics[scale=0.35, angle=0]{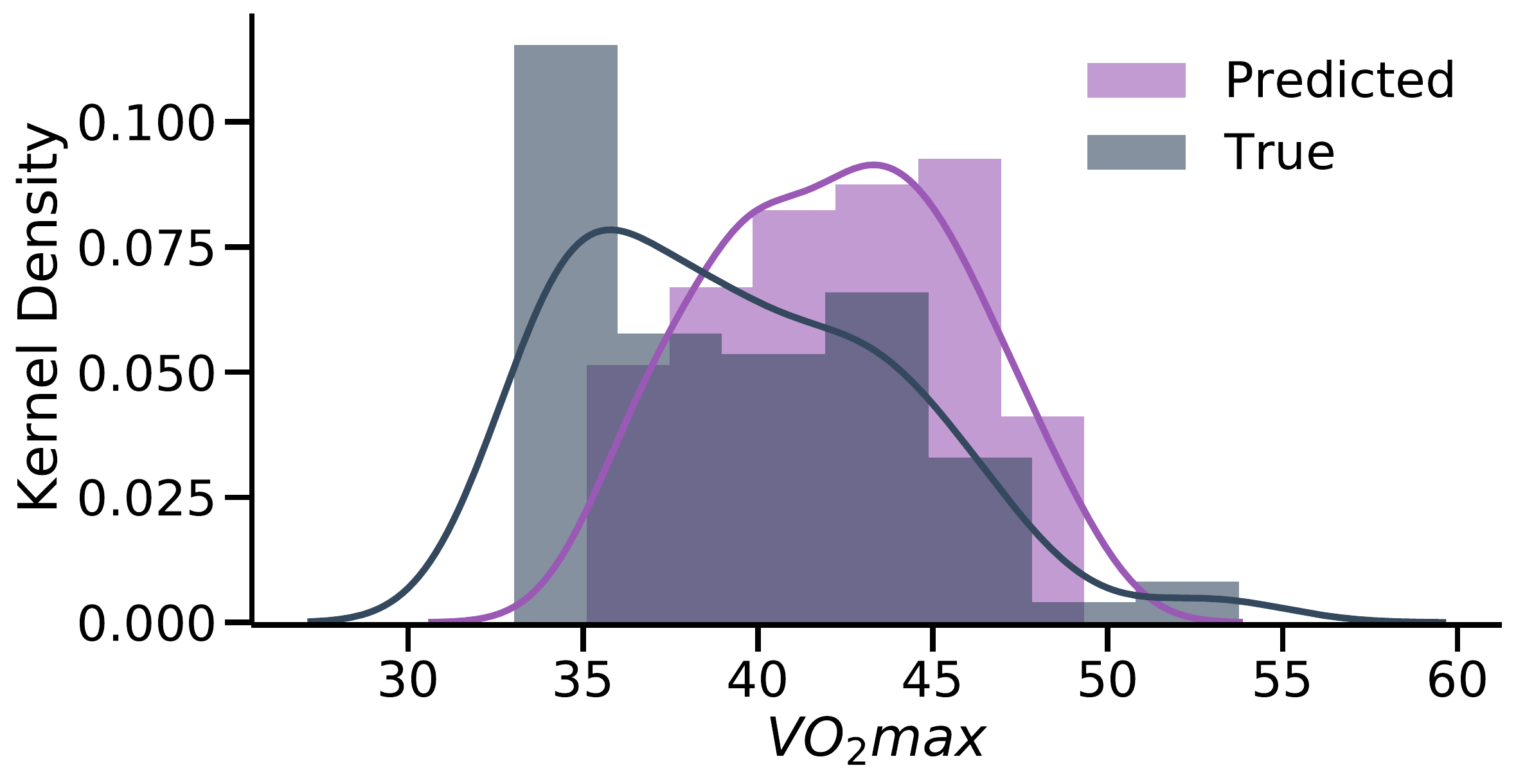}
        \caption{}
        \label{fig:BVS:b}
    \end{subfigure}
 
    \caption{\revised{\textbf{External validation of Fenland I model with maximal (peak exercise) test data using the BBVS cohort.} \textbf{(a)} Distribution of the predicted vs the true VO2max (RMSE=8.998) using all participants (N=181). \textbf{(b)} Distribution of the predicted vs the true VO2max (RMSE=5.19) by matching BBVS to have similar VO2max (mean$\pm$std) to the training set of Fenland I, using a subset of participants (N=82). Please see the main text for interpretation of these results.} } 
    \label{fig:BVS}
\end{figure}

\begin{figure*}[h]
\centering
\begin{subfigure}{0.45\textwidth}
\includegraphics[scale=0.45]{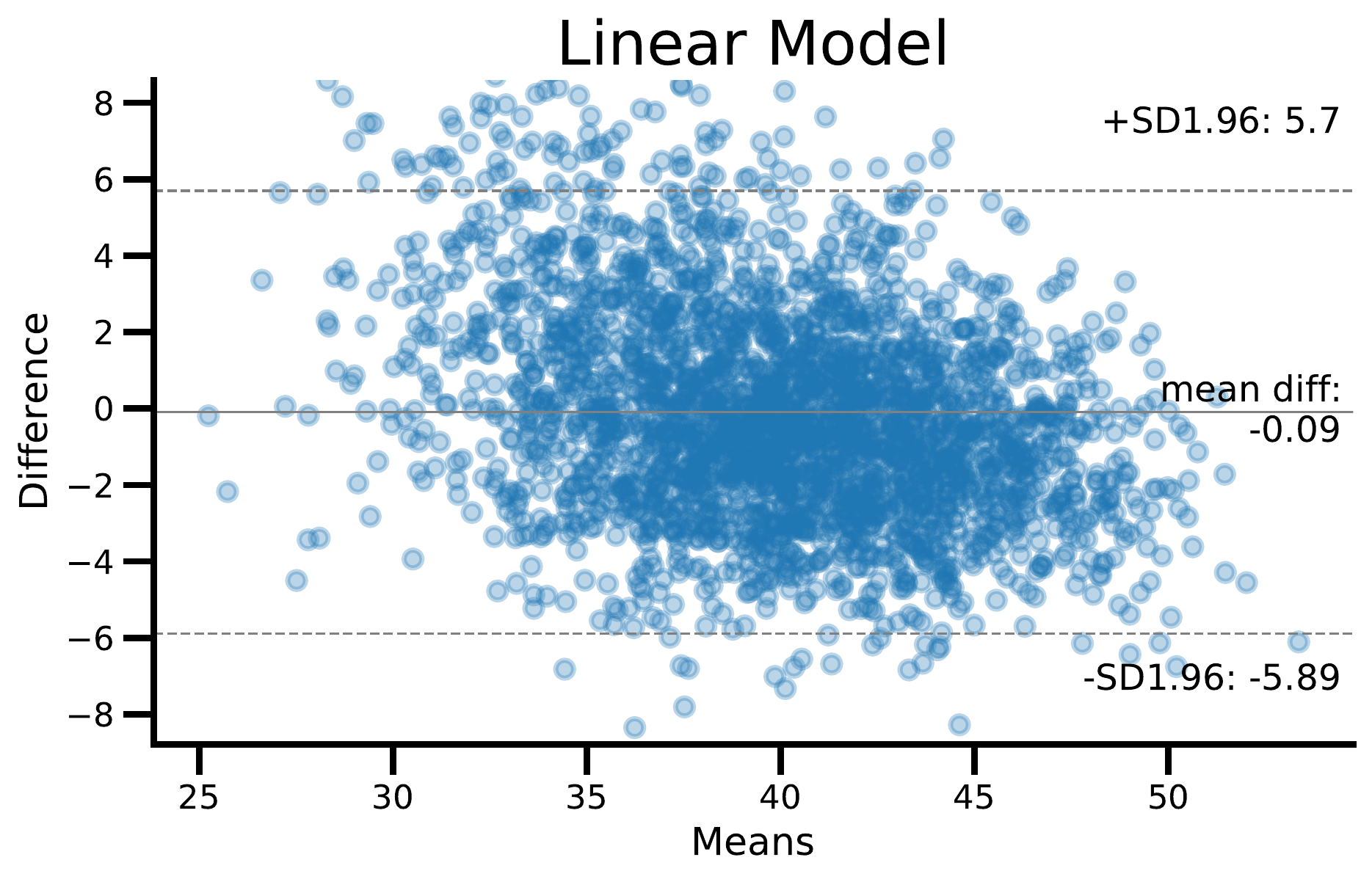}
\caption{Linear regression} \label{fig:bland_alt:a}
\end{subfigure}
\hspace{10pt}
\begin{subfigure}{0.45\textwidth}
\includegraphics[scale=0.45]{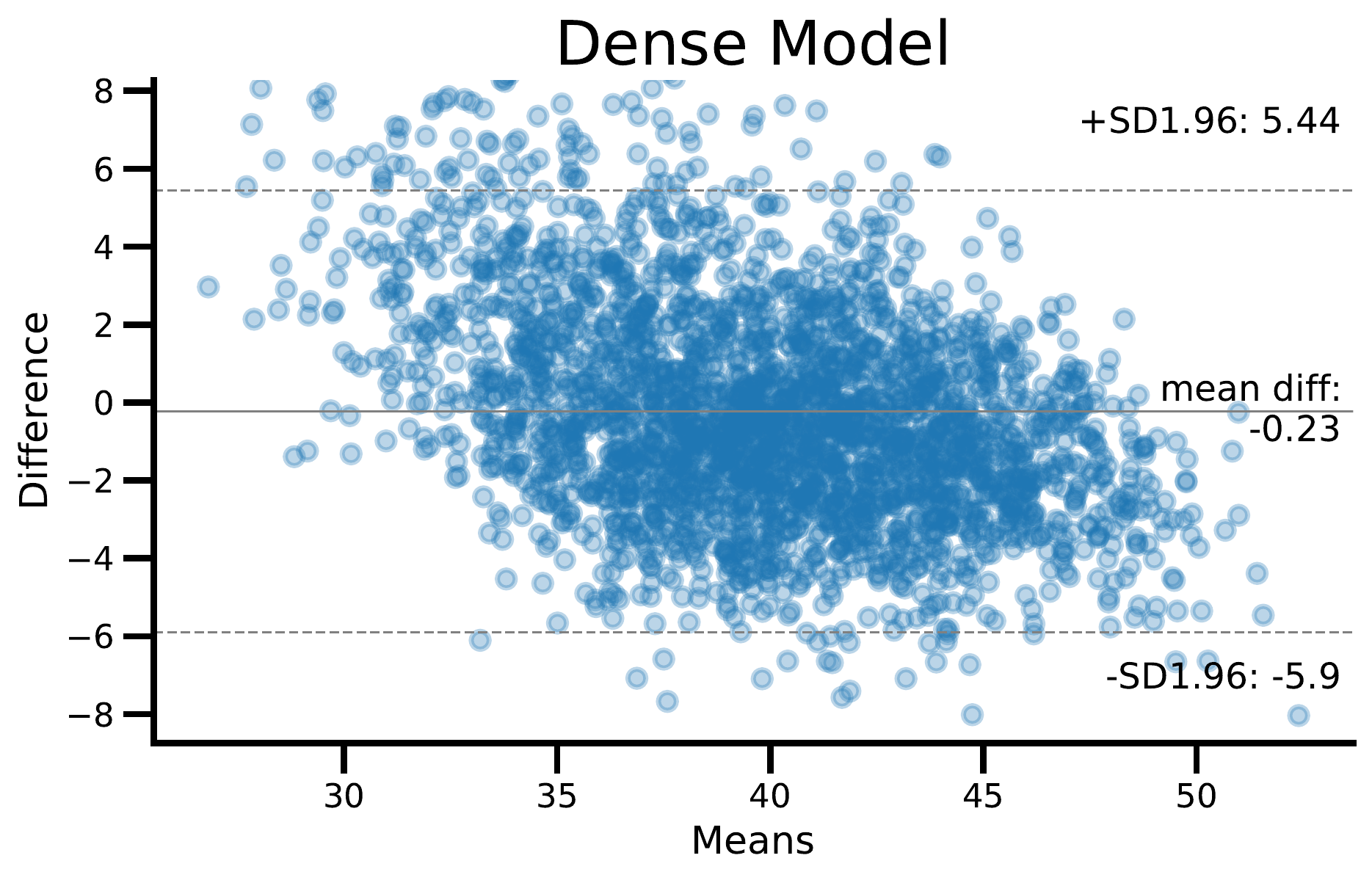}
\caption{Feature-based Dense neural network} \label{fig:bland_alt:b}
\end{subfigure}
 
    \caption{\npj{\textbf{Bland–Altman plots of the comprehensive linear and Dense models.} \textbf{(a-b)} Both models demonstrate low bias and high agreement between true and predicted values, with the Dense model showing lower standard deviation on the top values. 
     } } 
    \label{fig:bland_alt}
\end{figure*}

\begin{table*}
\caption{\textbf{Description of the features/variables used in our analysis as inputs to the models.} The features with asterisks(\textbf{*}) are time-series and therefore we have extracted the following statistical variables: \textit{mean, minimum, maximum, standard deviation, percentiles (25\%, 50\%, 75\%), and the slope of a linear regression fit}. The final set of features is 68.}
\begin{tabular}{@{}llll@{}}
\toprule
\multicolumn{3}{l}{\textbf{Features/Variables}}                                                                              & \textbf{Description}                                                                                                                                                                                                              \\ \midrule
\multicolumn{3}{l}{\textbf{Sensors}}                                                                                         &                                                                                                                                                                                                                                   \\
 & \multicolumn{2}{l}{Acceleration*}                                                                                         & Acceleration measured in m\textit{g}                                                                                                                                                                                              \\
 &                                                              &                                                            &                                                                                                                                                                                                                                   \\
 & \multicolumn{2}{l}{Heart rate (HR)*}                                                                                      & Mean HR resampled in 15sec intervals, measured in BPM                                                                                                                                                                             \\
 &                                                              &                                                            &                                                                                                                                                                                                                                   \\
 & \multicolumn{2}{l}{Heart Rate Variability (HRV)*}                                                                         & \begin{tabular}[c]{@{}l@{}}HRV calculated by differencing \\ the second-shortest and the second-longest inter-beat interval \\ (as seen in \cite{faurholt2017state}), measured in ms\end{tabular}                                                                                                                  \\
 &                                                              &                                                            &                                                                                                                                                                                                                                   \\
 & \multicolumn{2}{l}{\begin{tabular}[c]{@{}l@{}}Acceleration-derived \\ Euclidean Norm Minus One (ENMO)*\end{tabular}}      & \begin{tabular}[c]{@{}l@{}}ENMO-like variable (Acceleration/0.0060321) + 0.057) (as seen in  \cite{white2016estimation}) \end{tabular}                                                                                            \\
\multicolumn{2}{l}{}                                            &                                                            &                                                                                                                                                                                                                                   \\
 & \multicolumn{2}{l}{\begin{tabular}[c]{@{}l@{}}Acceleration-derived \\ Metabolic Equivalents of Task (METs)*\end{tabular}} &                                                                                                                                                                                                                                   \\
 &                                                              & Sedentary*                                                 & If Accelerometer \textless 1, take daily count and average                                                                                                                                                               \\
 &                                                              & Moderate to Vigorous*                                      & If Accelerometer \textgreater{}= 1,  take daily count and average                                                                                                                                                          \\
 &                                                              & Vigorous*                                                  & If Accelerometer \textgreater{}= 4.15, take daily count and average                                                                                                                                                       \\
\multicolumn{3}{l}{\textbf{Anthropometrics}}                                                                                 &                                                                                                                                                                                                                                   \\
 & \multicolumn{2}{l}{Age}                                                                                                   & Age, measured in years                                                                                                                                                                                                            \\
 & \multicolumn{2}{l}{}                                                                                                      &                                                                                                                                                                                                                                   \\
 & Sex                                                          &                                                            & Sex is binary (female/male)                                                                                                                                                                                                       \\
 & \multicolumn{2}{l}{}                                                                                                      &                                                                                                                                                                                                                                   \\
 & Weight                                                       &                                                            & Weight, measured in kilograms                                                                                                                                                                                                     \\
 & \multicolumn{2}{l}{}                                                                                                      &                                                                                                                                                                                                                                   \\
 & Height                                                       &                                                            & Height, measured in meters.centimeters                                                                                                                                                                                            \\
 & \multicolumn{2}{l}{}                                                                                                      &                                                                                                                                                                                                                                   \\
 & Body Mass Index (BMI)                                        &                                                            & BMI is calculated by Weight/($Heightˆ2$), measured in kg$/mˆ2$                                                                                                                                                                      \\
\multicolumn{3}{l}{\textbf{Resting Heart Rate}}                                                                              &                                                                                                                                                                                                                                   \\
 & \multicolumn{2}{l}{Wearable-derived RHR}                                                                                       & \begin{tabular}[c]{@{}l@{}}RHR is calculated by averaging the 4th, 5th, and 6th minute \\ of the baseline visit and adding to that the Sleeping Heart Rate\\ that has been inferred by the wearable device. \cite{gonzales2020resting} \end{tabular} \\
\multicolumn{3}{l}{\textbf{Seasonality}}                                                                                     &                                                                                                                                                                                                                                   \\
 & \multicolumn{2}{l}{Month of year}                                                                                         & \begin{tabular}[c]{@{}l@{}}The month number is used along with a coordinate encoding that \\ allows the models to make sense of their cyclical sequence.\end{tabular}                                                             \\ \bottomrule
  \label{tab:variables}
  \end{tabular}
\end{table*}


\end{document}